\documentclass[letterpaper]{article} 
\usepackage{aaai2026} 
\usepackage{times}  
\usepackage{helvet}  
\usepackage{courier}  
\usepackage[hyphens]{url}  
\usepackage{graphicx} 
\usepackage{natbib}  
\usepackage{caption} 
\usepackage{algorithm}
\usepackage{algorithmic}
\usepackage{amsmath} 

\usepackage{booktabs}
\usepackage{subcaption}

\usepackage{newfloat}
\usepackage{listings}
\DeclareCaptionStyle{ruled}{labelfont=normalfont,labelsep=colon,strut=off} 
\floatstyle{ruled}
\newfloat{listing}{tb}{lst}{}
\floatname{listing}{Listing}
\title{Uncertainty-Based Methods for Automated Process Reward Data Construction and Output Aggregation in Mathematical Reasoning}

\author {
    Jiuzhou Han\textsuperscript{\rm 1},
    Wray Buntine\textsuperscript{\rm 2},
    Ehsan Shareghi\textsuperscript{\rm 1}
}
\affiliations {
    \textsuperscript{\rm 1}Department of Data Science \& AI, Monash University\\
    \textsuperscript{\rm 2}College of Engineering and Computer Science, VinUniversity\\
    jiuzhou.han@monash.edu, wray.b@vinuni.edu.vn, ehsan.shareghi@monash.edu
}

\begin{document}

\maketitle

\begin{abstract}
Large language models have demonstrated remarkable capabilities in complex mathematical reasoning tasks, but they inevitably generate errors throughout multi-step solutions. Process-level Reward Models (PRMs) have shown great promise by providing supervision and evaluation at each intermediate step, thereby effectively improving the models’ reasoning abilities. However, training effective PRMs requires high-quality process reward data, yet existing methods for constructing such data are often labour-intensive or inefficient. In this paper, we propose an uncertainty-driven framework for automated process reward data construction, encompassing both data generation and annotation processes for PRMs. Additionally, we identify the limitations of both majority vote and PRMs, and introduce two generic uncertainty-aware output aggregation methods: Hybrid Majority Reward Vote and Weighted Reward Frequency  Vote, which combine the strengths of majority vote with PRMs. Extensive experiments on ProcessBench, MATH, and GSMPlus show the effectiveness and efficiency of the proposed PRM data construction framework, and demonstrate that the two output aggregation methods further improve the mathematical reasoning abilities across diverse PRMs. The code and data will be publicly available at \url{https://github.com/Jiuzhouh/UnPRM}.
\end{abstract}

\section{Introduction}
Inference-time scaling~\cite{DBLP:conf/iclr/LightmanKBEBLLS24} offers a practical approach to improving reasoning performance of Large Language Models (LLMs) by leveraging increased computational resources during inference. Within this framework, Process Reward Models (PRMs)~\cite{DBLP:journals/corr/abs-2211-14275} have been introduced to assess the correctness of intermediate reasoning steps, providing a fine-grained mechanism for scoring and filtering candidate solutions and thereby supporting more robust answer selection. Unlike Outcome Reward Models (ORMs)~\cite{DBLP:journals/corr/abs-2110-14168}, which focus solely on the final answer, PRMs provide fine-grained verification of the entire reasoning process. This granularity allows PRMs to identify solutions in which correct final answers are arrived at via flawed intermediate steps, revealing otherwise undetected errors in reasoning.

A major challenge in developing effective PRMs lies in the annotation of high-quality step-level supervision data, which is typically expensive and time-consuming. Early approaches~\cite{DBLP:conf/iclr/LightmanKBEBLLS24} relied on human annotation to ensure label quality, but this method is costly and not scalable. To improve efficiency, recent research has focused on automated annotation techniques, including Monte Carlo (MC) methods that estimate step correctness based on the probability of reaching the correct final answer~\cite{DBLP:conf/acl/WangLSXDLCWS24, DBLP:journals/corr/abs-2406-06592} and approaches that leverage strong LLMs as judges of step correctness~\cite{DBLP:journals/corr/abs-2502-11520, DBLP:journals/corr/abs-2406-14024}. However, these methods can require substantial computational resources and often suffer from inefficiency.

To overcome these challenges, we propose an uncertainty-driven PRM data construction framework. Our approach first generates candidate solutions for annotation using uncertainty-guided sampling, then applies an uncertainty-driven automated annotation process to efficiently and accurately label the correctness of each reasoning step. This pipeline significantly improves both the quality and scalability of PRM training data.

While PRMs and Majority Vote~\cite{DBLP:conf/iclr/0002WSLCNCZ23} are commonly used for aggregating LLM outputs, both have notable limitations. Majority Vote, which selects the most frequent answer among sampled outputs, can fail when answers are highly dispersed or when the model confidently produces incorrect solutions, leading to erroneous consensus. Conversely, PRMs may select suboptimal answers when faced with out-of-distribution or particularly challenging problems. To address these limitations and improve output aggregation, we further propose two uncertainty-aware hybrid strategies: Hybrid Majority Reward (HMR) Vote and Weighted Reward-Frequency (WRF) Vote. These methods combine the implicit confidence signals of Majority Vote with the explicit, step-level feedback from PRMs, aiming to achieve more reliable answer selection.

We construct several variations of PRM training data and conduct extensive experiments on ProcessBench~\cite{DBLP:journals/corr/abs-2412-06559}, MATH~\cite{DBLP:conf/nips/HendrycksBKABTS21}, and GSMPlus~\cite{DBLP:conf/acl/LiCZKB24} to validate the efficiency and effectiveness of our uncertainty-driven PRM data construction framework. Additionally, we demonstrate that our proposed uncertainty-aware aggregation strategies generalise well across different PRMs and yield notable performance improvements compared to majority vote and traditional PRM methods, with WRF generally proving to be the more robust strategy.


\section{Related Work}

\subsection{PRMs in Mathematical Reasoning}
Mathematical reasoning tasks are among the most challenging for LLMs, as they demand rigorous multi-step logical thinking and precise manipulation of mathematical symbols, leaving little room for ambiguity or guesswork. A single error at any step can invalidate the entire solution, and the abstract, concise nature of mathematical language further compounds the difficulty. 
To assess the correctness of these intermediate steps, PRMs~\cite{DBLP:conf/iclr/LightmanKBEBLLS24, DBLP:journals/corr/abs-2211-14275} have been introduced. PRMs assign a score to each reasoning step, reflecting its likelihood of correctness. By aggregating these step scores, a final score for the entire solution can be obtained, providing an overall assessment of its validity. Furthermore, PRMs enable the selection of the highest-scoring solution from multiple candidates, thereby further improving the reasoning performance of LLMs.

\subsection{Mathematical Reasoning Step Verification}
Although PRMs demonstrate significant potential for enhancing the performance of LLMs, their effectiveness relies heavily on the availability of high-quality labelled training data, particularly for evaluating the correctness of intermediate reasoning steps. There are three primary approaches to data annotation: (1) Human annotation~\cite{DBLP:conf/iclr/LightmanKBEBLLS24}, which can yield highly accurate labels but is labour-intensive, expensive, and time-consuming; (2) LLM-as-a-Judge methods~\cite{DBLP:journals/corr/abs-2502-11520, DBLP:conf/iclr/ZhangHBKKA25, DBLP:journals/corr/abs-2406-14024}, which prompt LLMs to directly assess step correctness, offering automation but incurring high computational costs, especially when labelling large datasets, and inheriting the limitations of LLMs such as hallucination and inaccuracies; and (3) inferring step correctness from solution outcomes using techniques such as Monte Carlo estimation. The latter enables automated annotation by leveraging open-source LLMs~\cite{DBLP:conf/acl/WangLSXDLCWS24}, but often suffers from inefficiency. To address this, \citet{DBLP:journals/corr/abs-2406-06592} introduced a more efficient Monte Carlo Tree Search algorithm using binary search, while \citet{DBLP:journals/corr/abs-2503-02382} further improved efficiency with an adaptive binary search method. Building on these developments, we propose an uncertainty-driven search algorithm for automated step label annotation, further enhancing annotation efficiency without sacrificing quality.

\subsection{Uncertainty Estimation in LLMs}

Uncertainty estimation plays a crucial role in assessing the confidence of outputs generated by LLMs. Logits-based methods~\cite{DBLP:conf/acl/HanBS24, DBLP:journals/corr/abs-2310-04782} quantify uncertainty by analysing the token-level logits produced by the model; however, these approaches are only feasible when direct access to output token logits is available. Alternatively, self-consistency~\cite{DBLP:conf/iclr/0002WSLCNCZ23} provides an implicit measure of model confidence by sampling multiple outputs for a given question—answers that appear frequently among the samples are considered high-confidence, while those with low frequency indicate greater uncertainty. Prior studies~\cite{DBLP:conf/acl/HanBS24, DBLP:journals/corr/abs-2505-19590} have found a strong correlation between higher model confidence and answer correctness, with errors being more likely when the model is uncertain. Motivated by these findings, we propose leveraging uncertainty as a guiding signal to identify erroneous steps in mathematical reasoning solutions, thereby improving the automated annotation process for PRM training data.

\section{Uncertainty-driven Automated Process Reward Data Construction}
In this section, we first introduce the uncertainty estimation method employed in our approach. We then describe how this uncertainty estimation is leveraged to guide PRM data generation, and finally, we detail the design of our uncertainty-driven automated PRM data annotation process.

\subsection{Uncertainty Estimation}

To quantify the uncertainty associated with each candidate solution generated by the language model, we employ an entropy-based uncertainty estimation approach. Specifically, for a given solution consisting of a sequence of $n$ tokens, we extract the log-probability assigned by the model to each generated token during the decoding process. Let $p_i$ denote the token log-probability assigned to the $i$-th token in the solution, then we apply a softmax function to the token log-probabilities to obtain the probabilities $[z_1, z_2, ..., z_n]$. The overall uncertainty $u$ of the candidate solution is then computed as the entropy over the sequence of token probabilities, which captures the model’s average uncertainty throughout the solution. Formally, the uncertainty score is calculated as $u = -\sum_{i=1}^{n} z_i \cdot \log(z_i)$, where higher values of $u$ indicate that the model was generally less confident across the sequence, while lower values indicate more confident and deterministic predictions. This entropy-based uncertainty metric
supports our uncertainty-driven PRM data generation and automated step-level annotation.


\subsection{Uncertainty-driven PRM Data Generation Process}

To construct the training data for PRMs, 
given a set of questions, each paired with its ground-truth answer, we first sample $k$ candidate solutions using an LLM, obtaining token-level log-probabilities for each generated solution. For each candidate solution, we compute an uncertainty score based on its token log-probabilities using an uncertainty function. We then verify each solution using the ground-truth final answer to categorise it as correct or incorrect. Among all correct solutions, we select the top $m$ with the \emph{highest} uncertainty scores, and similarly, among incorrect solutions, we select the top $n$ most uncertain. The final candidate set comprises both the most uncertain correct and incorrect solutions, ensuring diversity and difficulty in the PRM training data. This targeted sampling encourages the PRM to learn more robustly from ambiguous or challenging reasoning trajectories, thereby improving its ability to identify and discriminate step-level correctness during inference. Full algorithm is outlined in technical appendix.

\subsection{Uncertainty-driven PRM Data Annotation Process}
To efficiently annotate the step-level correctness of candidate solutions, we propose an automated uncertainty-driven step label annotation process (Algorithm~\ref{alg:uncertainty_process_annotation}). Given a set of candidate solutions $C$ partitioned into correct and incorrect solutions, we first assign \texttt{True} (correct) labels to all steps within each correct solution (LN3-7), assuming there are no mistakes in the intermediate steps. For a given incorrect solution composed of $T$ steps, we compute the uncertainty $u(s_t)$ at each individual step $s_t$, and subsequently calculate the uncertainty delta $\Delta u(s_t) = u(s_t) - u(s_{t-1})$ for steps $t = 2, \ldots, T$ (LN8-11). These deltas serve to identify steps where the model’s uncertainty exhibits the greatest increase, which are likely to correspond to points where reasoning errors occur (see Table~\ref{tab:data_stat} for supporting evidence). Steps are then ordered by descending uncertainty delta and placed in a list to prioritise annotation at the most uncertain transitions.

For each candidate step in this ordered list, we employ an adaptive sampling strategy following the prior work~\cite{DBLP:journals/corr/abs-2503-02382}. Starting from the selected step $s_i$, we sample $N$ new solution completions (LN12-18).\footnote{The sample size $N$ is dynamically increased until a minimum number of correct samples ($N_\mathrm{min}$) is reached, or until a maximum sampling threshold ($N_\mathrm{max}$) is met.
The adaptive sampling strategy reduces computational overhead while ensuring annotation quality.} We refer to the set containing all these solutions for $s_i$ as $S^i_{\text{all}}$ and the set containing only the solutions ending with correct answer as $S^i_{\text{correct}}$. We refer to solution trajectory $k$ in these sets as $\text{traj}^k$ and its perplexity as $PPL(\text{traj}^k)$.\footnote{Perplexity of a sequence $X$ of length $L$ is calculated as $\exp \left( -\frac{1}{L} \sum_{l=1}^{L} \log p(x_l \mid x_{<l}) \right).$} The Monte Carlo-based perplexity $MC_{PPL}$~\cite{DBLP:journals/corr/abs-2503-02382} is computed as,
\begin{equation} 
MC_{PPL}(s_i)= \frac{\sum_{\text{traj}^k\in S^i_{\text{correct}}} \log PPL(\text{traj}^k)}{\sum_{\text{traj}^k\in S^i_{\text{all}}} \log PPL(\text{traj}^k)}
\end{equation}

A threshold $\tau$ is defined as the MC perplexity corresponding to the initial problem state, where the input consists solely of the question. If the $MC_{PPL}$ of the step $s_i$ falls below the threshold $\tau$, steps up to $s_{i-1}$ are labelled as \texttt{True} (correct), and all subsequent steps from $s_i$ onwards are labelled as \texttt{False} (incorrect) (LN19-23). In contrast to approaches that identify the first error step, our automated PRM data annotation process is designed to locate the most uncertain reasoning error within each incorrect solution. The annotated solutions, each with step-level correctness labels, are then aggregated to form the final training set $\widetilde{C}$.

\begin{algorithm}[tb]
\caption{Uncertainty-driven Step Label Annotation}
\label{alg:uncertainty_process_annotation}
\textbf{Input:} Solution set $C$; uncertainty function $u(\cdot)$\\
\textbf{Param:} $N_0$, $N_\mathrm{min}$, $N_\mathrm{max}$, $\tau$\\
\textbf{Output:} Labelled solution set $\widetilde{C}$

\begin{algorithmic}[1]
\STATE $\widetilde{C} \gets \emptyset$
\FOR{$s \in C$}
   \STATE $U \gets \emptyset$
    \IF{$s \in C_\mathrm{correct}$}
        \STATE Label all steps in $s$ as \texttt{True}
        \STATE Append labelled $s$ to $\widetilde{C}$; \textbf{continue}
    \ENDIF
    \FOR{$s_t \in s, t > 1$} 
        \STATE Compute $\Delta u(s_t) = u(s_t) - u(s_{t-1})$
        \STATE Append $(\Delta u(s_t), s_t)$ to $U$
    \ENDFOR
    \STATE Sort $U$ in desc. order according to $\Delta u$
    \FOR{$s_i$ in $U$}
        \STATE $N \gets N_0$
        \REPEAT
            \STATE Sample $N$ solutions from $s_i$; count $N_\mathrm{correct}$
            \STATE Increase $N$ dynamically if $N_\mathrm{correct} < N_\mathrm{min}$
        \UNTIL{$N_\mathrm{correct} \geq N_\mathrm{min}$ or reaching $N_\mathrm{max}$}
        \STATE Compute $MC_{PPL}$ over samples using Eq.~(1)
        \IF{$MC_{PPL} < \tau$}
            \STATE Label $[s_1..s_{i-1}]$ as \texttt{True}, $[s_i..s_T]$ as \texttt{False}
            \STATE Append labelled $s$ to $\widetilde{C}$; \textbf{break}
        \ENDIF
    \ENDFOR
\ENDFOR
\STATE \textbf{return} $\widetilde{C}$
\end{algorithmic}
\end{algorithm}

\section{Uncertainty-aware Output Aggregation}

In this section, we first discuss the limitations of commonly used output aggregation methods, such as Majority Vote and PRM-based approaches. We then propose two aggregation strategies that combine the strengths of both Majority Vote and PRMs to achieve more robust answer selection.

\subsection{Limitations of Majority Vote and PRMs}

LLMs often generate diverse solutions when prompted multiple times for the same problem, due to their inherent stochasticity and the complexity of mathematical reasoning tasks. Majority Vote~\cite{DBLP:conf/iclr/0002WSLCNCZ23} and PRMs~\cite{DBLP:conf/iclr/LightmanKBEBLLS24} are two commonly used and effective approaches to aggregate these multiple candidate outputs and identify the most likely correct answer from LLMs. 

In Majority Vote, the answer that occurs most frequently among the sampled solutions is chosen, implicitly favouring the response that the model generates with the highest confidence. While this method is straightforward and often effective, it can fail when the candidate answers are highly dispersed or when the model is consistently confident in an incorrect solution, leading to erroneous consensus.

In contrast, PRMs provide a more explicit evaluation by predicting the correctness of each reasoning step within a solution, offering step-level signals to inform the answer selection process. This approach can more robustly identify subtle reasoning errors that majority vote may overlook. However, the effectiveness of PRM-based methods depends on the reward model’s ability to accurately assess a wide range of problem types; inaccuracies in the reward model can result in the selection of sub-optimal answers, especially for out-of-distribution or particularly challenging examples.

Given these complementary strengths and limitations, we propose two hybrid strategies: Hybrid Majority Reward (HMR) Vote and Weighted Reward-Frequency (WRF) Vote, which are designed to integrate the implicit confidence signals from majority vote with the explicit feedback signals provided by PRMs. By leveraging both consensus among candidate solutions and fine-grained reasoning evaluation, these methods aim to achieve more reliable and accurate answer aggregation for LLM-generated outputs.

\subsection{Hybrid Majority Reward Vote}

In Hybrid Majority Reward (HMR) vote strategy, given a set of $N$ sampled candidate solutions, each comprising a sequence of reasoning steps and a final answer, the HMR method first extracts the answer from each solution to form a collection of candidate answers. Then, the most frequent answer through majority vote is determined. If the majority answer appears in at least half of the solutions ($f_{\mathrm{maj}} \geq N/2$), it is directly selected as the final answer. However, if no answer achieves a strict majority ($f_{\mathrm{maj}} < N/2$), indicating ambiguity or lack of consensus among candidates, the algorithm leverages the PRM to guide answer selection. Specifically, for each candidate solution, the PRM function is invoked to compute a list of step-wise scores, and the minimum score across the steps is used as the overall reward for this solution. The answer given by the solution with the highest reward is selected as the final answer. This hybrid strategy combines the robustness of majority vote with the fine-grained reasoning assessment provided by the PRM, ensuring both consensus and confidence inform the answer aggregation process.

\subsection{Weighted Reward Frequency Vote}

Weighted Reward-Frequency (WRF) vote strategy provides another answer aggregation method. Given a set of $N$ sampled candidate solutions, each consisting of a series of reasoning steps and a final answer, the WRF algorithm integrates both the solution’s frequency and the quality of its reasoning steps as evaluated by the PRM. For each candidate solution, the PRM function computes step-wise scores, and the minimum score is taken as the overall reward for the solution. These rewards are grouped according to their corresponding final answers.

For each unique answer, the algorithm calculates the mean PRM reward and the frequency with which that answer appears among the candidates. Both metrics are then individually normalised across all answers to ensure comparability. Specifically, the normalised mean reward $\hat{m}_a$ and the normalised frequency $\hat{f}_a$ for answer $a$ are computed using min-max normalisation. The final combined score for each answer is calculated as a weighted sum of its normalised mean reward and frequency, controlled by the weighting parameter $\alpha \in [0, 1]$. The answer with the highest combined score is selected as the output.

By integrating both the consensus among candidate solutions (frequency) and the confidence derived from step-level PRM rewards, the WRF vote approach offers a more nuanced and fine-grained mechanism for answer aggregation. In our experiments, we set $\alpha = 0.5$, giving equal weight to both frequency and reward components.

\section{Experiments}

\subsection{PRM Training Data Construction}

We utilise the MATH dataset~\cite{DBLP:conf/nips/HendrycksBKABTS21}, which comprises 12,500 challenging competition-level mathematics problems, to construct our PRM training data. Following the EpicPRM~\cite{DBLP:journals/corr/abs-2503-02382} protocol, we adopt the same 3,500 randomly selected questions from the MATH training set as our seed questions. To enhance the diversity of sampled Chain-of-Thought (CoT) reasoning solutions for each question, we employ three different LLMs: Llama-3.1-8B-Instruct~\cite{DBLP:journals/corr/abs-2407-21783}, Qwen2.5-7B-Instruct~\cite{DBLP:journals/corr/abs-2412-15115}, and Mistral-7B-Instruct-v0.3~\cite{DBLP:journals/corr/abs-2310-06825}. For each model, we set the sampling temperature to 0.8 and generate 32 solutions per mathematics question. Subsequently, we apply the uncertainty-driven PRM data generation method to select the 2 most uncertain correct solutions and 6 most uncertain incorrect solutions.

We then filter out solutions with undesired formats (e.g., answers not presented in the correct format) and split the remaining candidate solutions into intermediate steps for automated annotation. For step-level annotation, we employ our automated uncertainty-driven step label annotation method, which assigns True or False labels to each step of the candidate solutions from the three LLMs. This process yields $40K$ labelled training examples, referred to as UnPRM40K.

For comparison, prior work~\cite{DBLP:journals/corr/abs-2503-02382} uses similarity, rather than uncertainty, as the selection criterion. Specifically, it selects candidate solutions exhibiting the lowest cosine similarity scores. We follow this approach to generate candidate solutions and annotate them using the uncertainty-driven step label annotation method, resulting in another $40K$ training examples, denoted as SimPRM40K.

The uncertainty-driven step label annotation method identifies the most uncertain error step in the solution. To compare this with an alternative approach that labels data based on the first error step, we re-annotate the same $40K$ examples using the adaptive binary search method adopted in EpicPRM~\cite{DBLP:journals/corr/abs-2503-02382}, referred to as EpicPRM40K.

To further investigate the effect of the error step’s location on model performance, we conduct an additional experiment in which the error step for incorrect candidate solutions is selected at random, denoted as RanPRM40K.

\subsection{PRM Training}

The objective of PRMs is to determine, at each step of the solution process, whether the reasoning trajectory remains correct. Given a problem $q$ and a sequence of solution steps $s_1 \rightarrow s_2 \rightarrow \cdots \rightarrow s_t$, the PRM model assigns a score $y_t$ between 0 and 1 to indicate the correctness of the step. This formulation naturally leads to a binary classification framework, where the model outputs a probability reflecting whether the solution is still correct up to the current step.

To train the PRM, we employ supervised fine-tuning of a language model. The input consists of the problem statement concatenated with the intermediate reasoning steps, each separated by a special step tag (in our case, the Unicode character `ş'). This tag is inserted between each step to delineate step boundaries within the input sequence. For every step, the label is either correct (`+') or incorrect (`-'), and these labels are aligned with the tokens immediately following each step tag in the input sequence. This explicit tagging enables the model to attend to the step boundaries and precisely associate each predicted label with its corresponding reasoning step.

The model is optimised using the binary cross-entropy loss with logits, targeting the prediction of the correct token at each annotated step. The loss function used for training is given by:
\begin{equation}
\mathcal{L} = -\frac{1}{N} \sum_{i=1}^{N} \left[\, y_i \log \hat{p}_i + (1 - y_i) \log(1 - \hat{p}_i)\, \right]
\end{equation}
where $N$ is the number of step-level predictions in a batch, $y_i \in \{0, 1\}$ is the ground-truth label for the $i$-th step (1 for correct, 0 for incorrect), and $\hat{p}_i$ is the model’s predicted probability for the correctness of step $i$, obtained by applying the sigmoid function to the output logits.

During inference, the PRM predicts a step score for each intermediate step by extracting the logits associated with the candidate tokens at each step tag position and applying a softmax or sigmoid function. The resulting probability assigned to the correct token (`+') at each step reflects the model’s confidence in the correctness of the current reasoning process. This allows the PRM to provide fine-grained scores for each step.

We use Qwen2.5-Math-7B-Instruct~\cite{DBLP:journals/corr/abs-2409-12122} as the base model for PRM training. The PRMs are fine-tuned for three epochs on a single A100 GPU using the LoRA~\cite{DBLP:conf/iclr/HuSWALWWC22} technique for parameter-efficient adaptation.

\subsection{PRM Best-of-N Evaluation}

Best-of-N (BoN) evaluation, which selects the highest-scored solution from N candidates according to the PRM, is a widely-used approach for assessing PRM performance in prior research~\cite{DBLP:conf/iclr/LightmanKBEBLLS24, DBLP:conf/acl/WangLSXDLCWS24, DBLP:journals/corr/abs-2501-07301}. Following this approach, we sample up-to 128 candidate solutions for each problem, and use the minimum step score as the overall solution score.

\subsubsection{Datasets}
We evaluate the PRMs on two mathematical reasoning datasets: MATH~\cite{DBLP:conf/nips/HendrycksBKABTS21} and GSMPlus~\cite{DBLP:conf/acl/LiCZKB24}. For MATH, we use the same test set as in prior work~\cite{DBLP:conf/iclr/LightmanKBEBLLS24}, which consists of 500 math problems uniformly sampled at random from all categories. GSMPlus is an augmented version of GSM8K~\cite{DBLP:journals/corr/abs-2110-14168} that introduces various mathematical perturbations, enabling a comprehensive assessment of LLMs’ robustness in mathematical reasoning. As GSMPlus comprises eight types of questions, we uniformly select 50 questions from each type, resulting in 400 problems.
\begin{figure}[t]
  \centering
  \begin{subfigure}[b]{0.23\textwidth}
    \includegraphics[width=\textwidth]{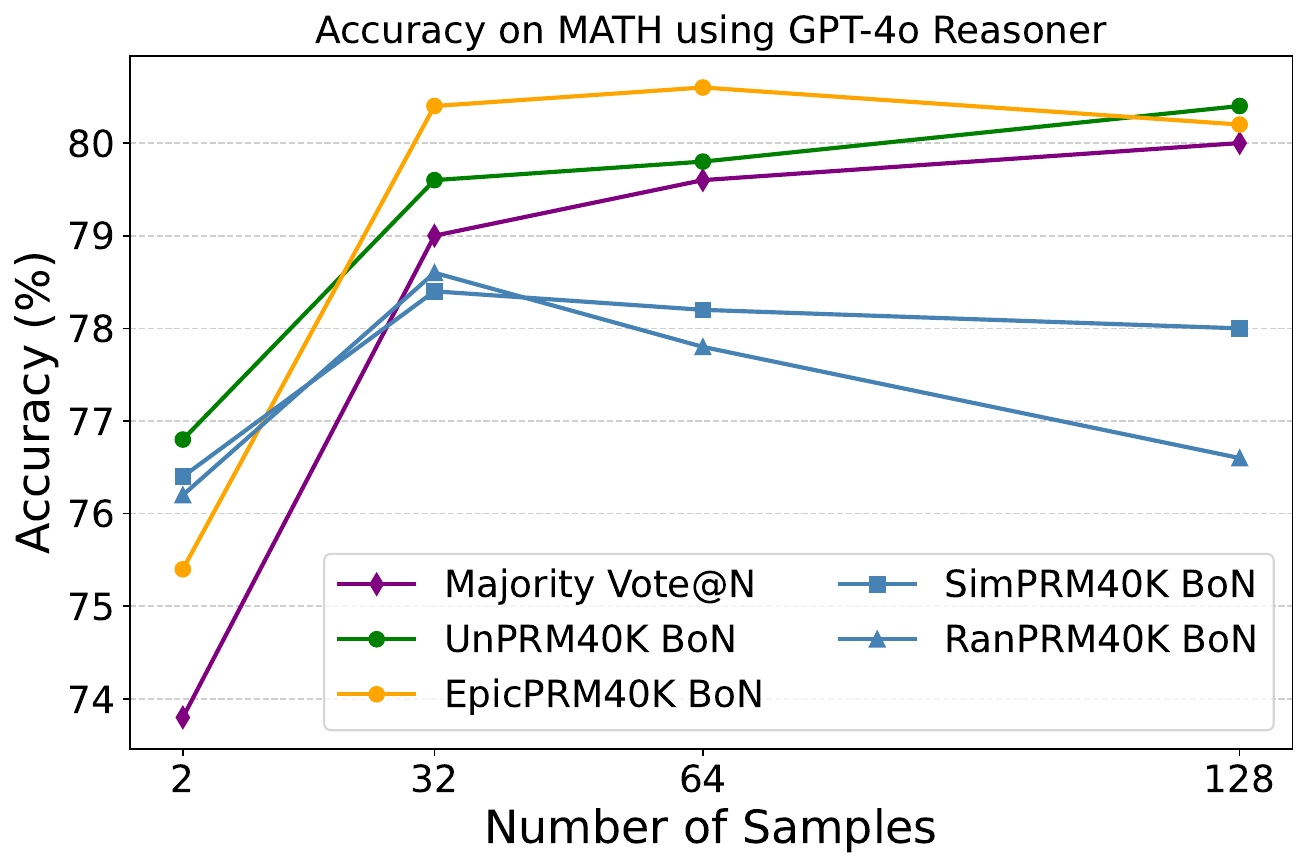}
  \end{subfigure}
  \begin{subfigure}[b]{0.23\textwidth}
    \includegraphics[width=\textwidth]{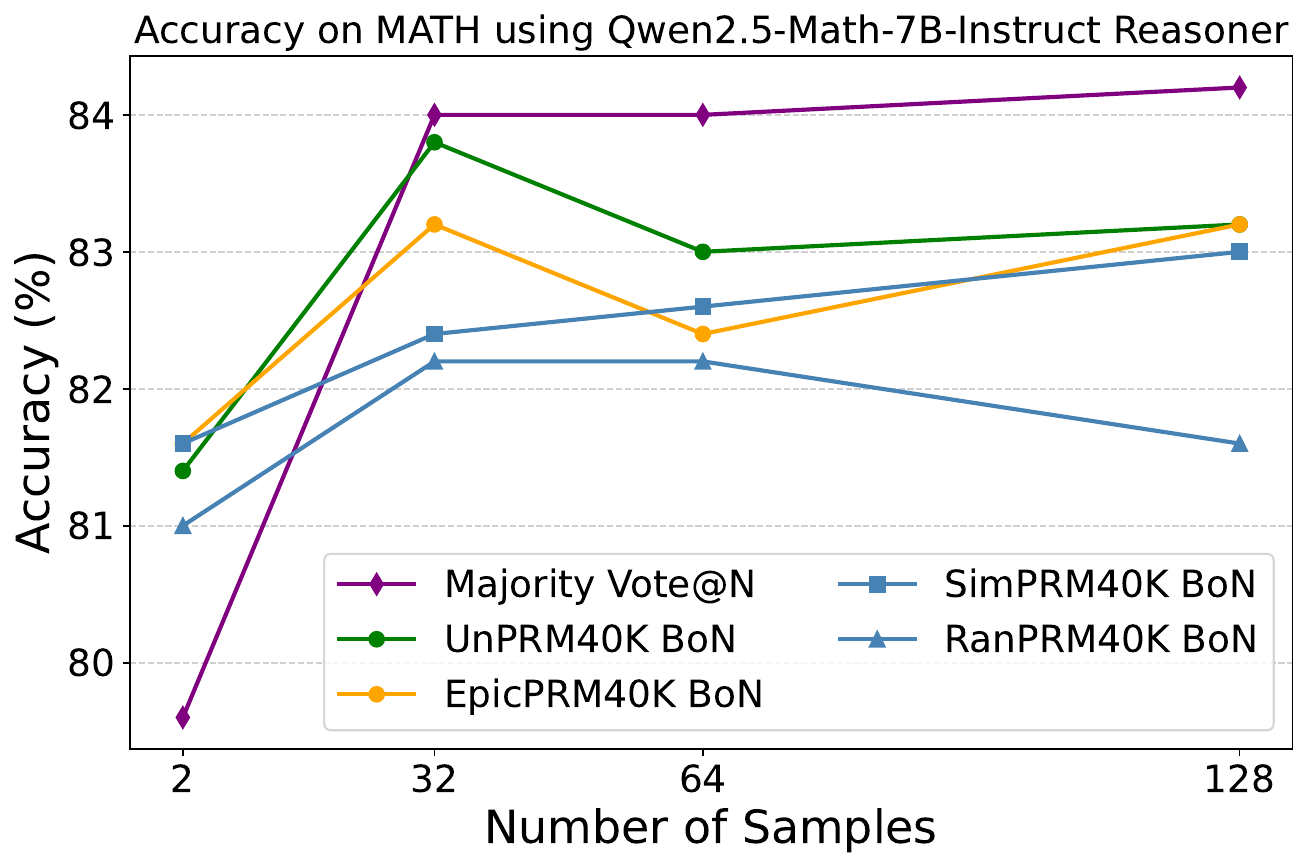}
  \end{subfigure}
  \\ 
  \begin{subfigure}[b]{0.23\textwidth}
    \includegraphics[width=\textwidth]{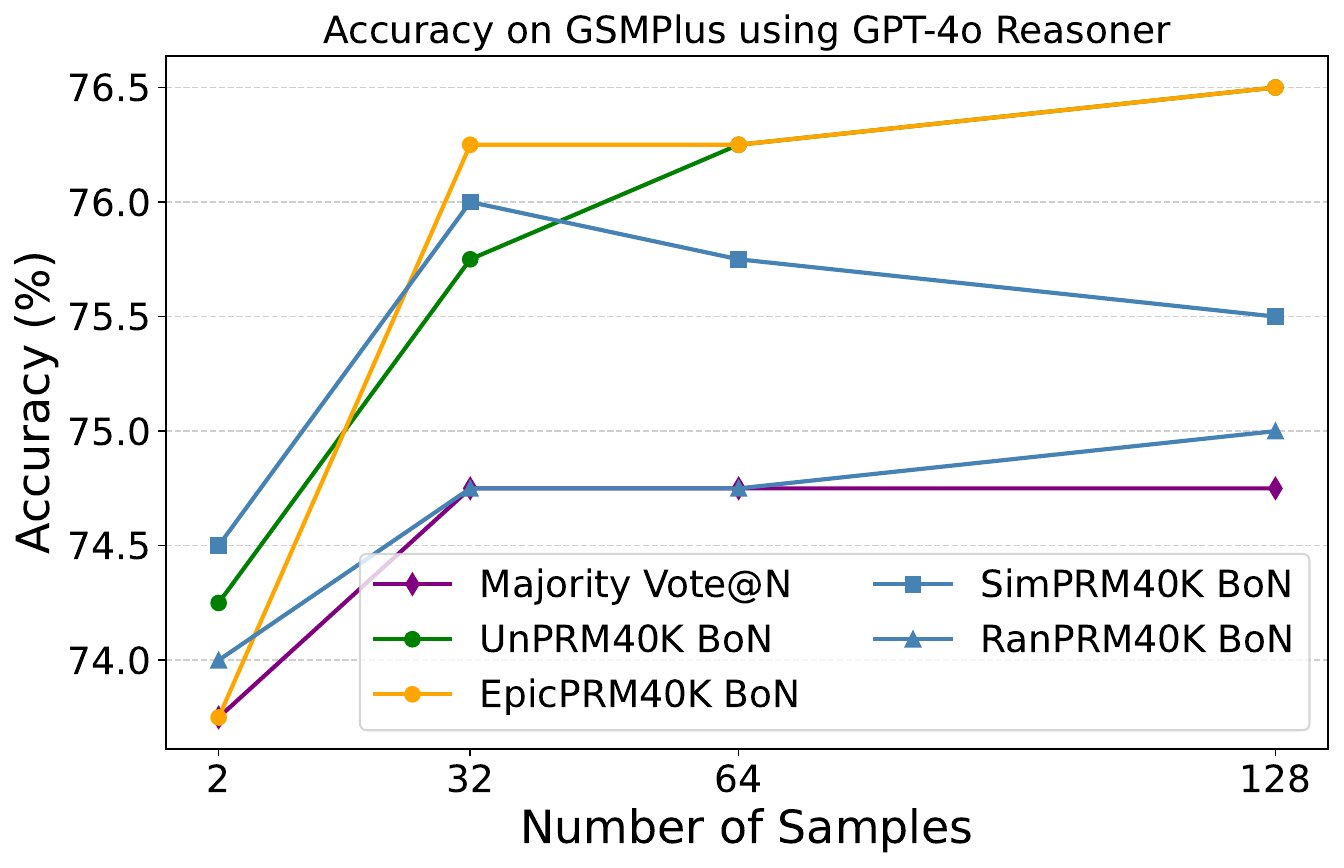}
  \end{subfigure}
  \begin{subfigure}[b]{0.23\textwidth}
    \includegraphics[width=\textwidth]{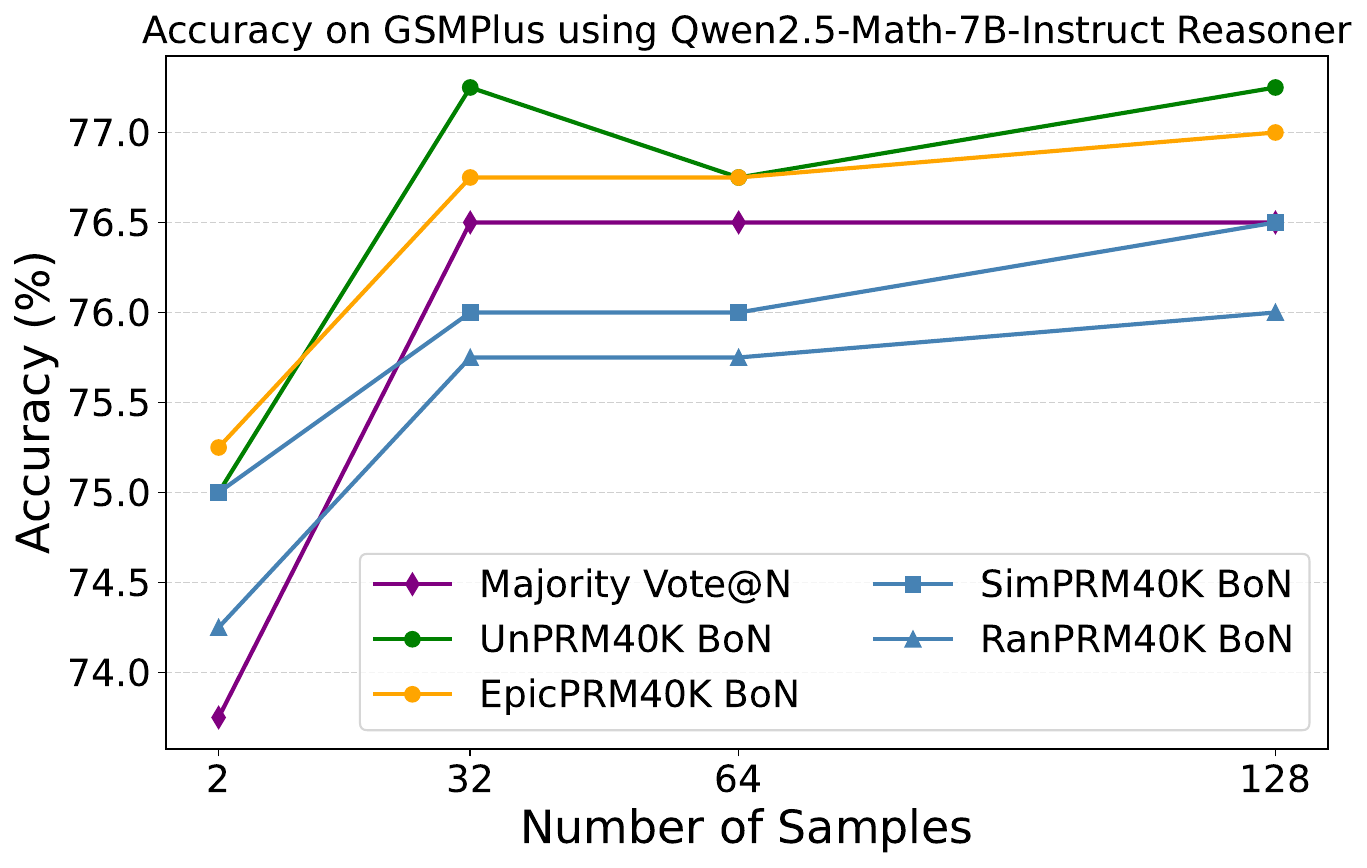}
  \end{subfigure}
  \caption{Evaluation results of different PRMs with BoN strategies on two datasets using two reasoners.}
  \label{fig:40kprms_results}
\end{figure}
\begin{figure*}[t]
  \centering
  \begin{subfigure}[b]{0.3\textwidth}
    \includegraphics[width=\textwidth]{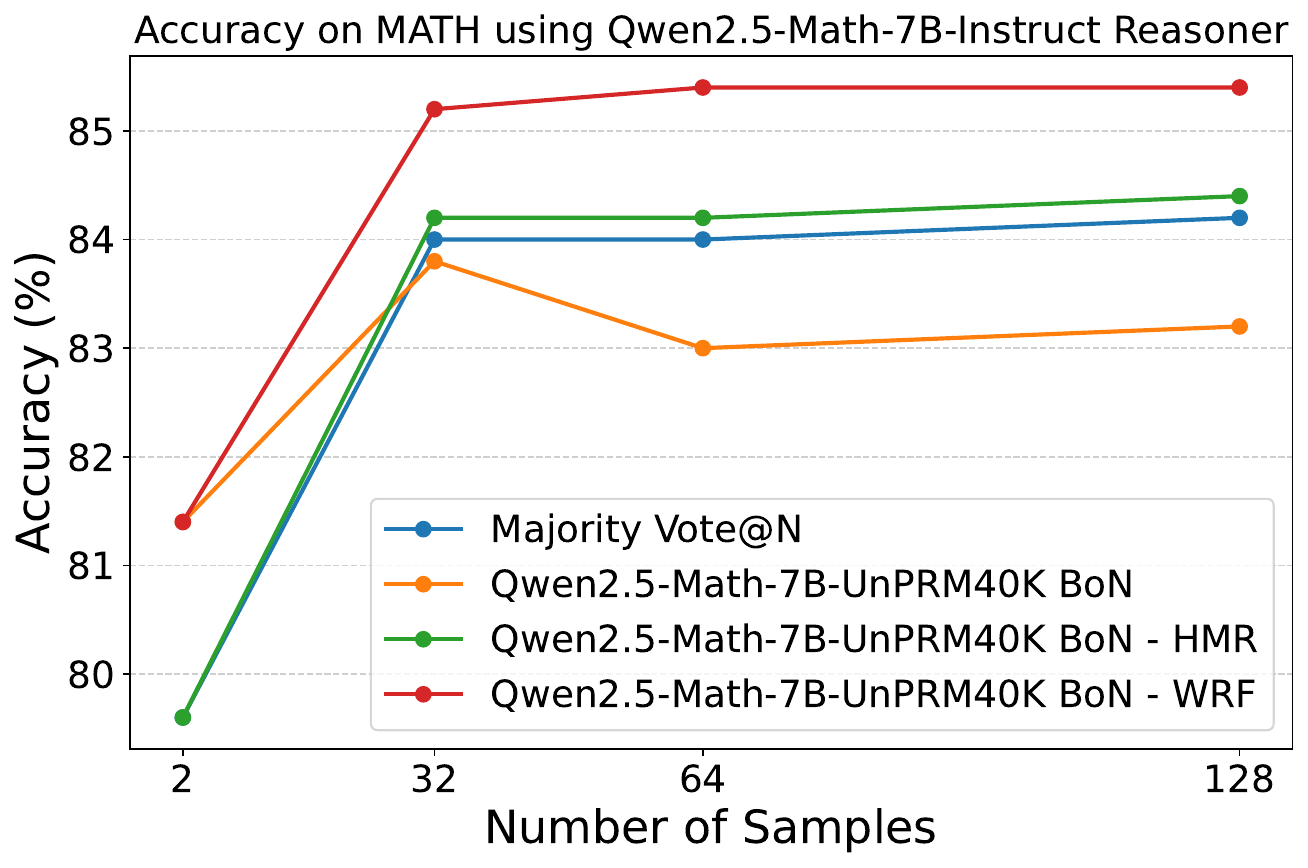}
  \end{subfigure}
  \begin{subfigure}[b]{0.3\textwidth}
    \includegraphics[width=\textwidth]{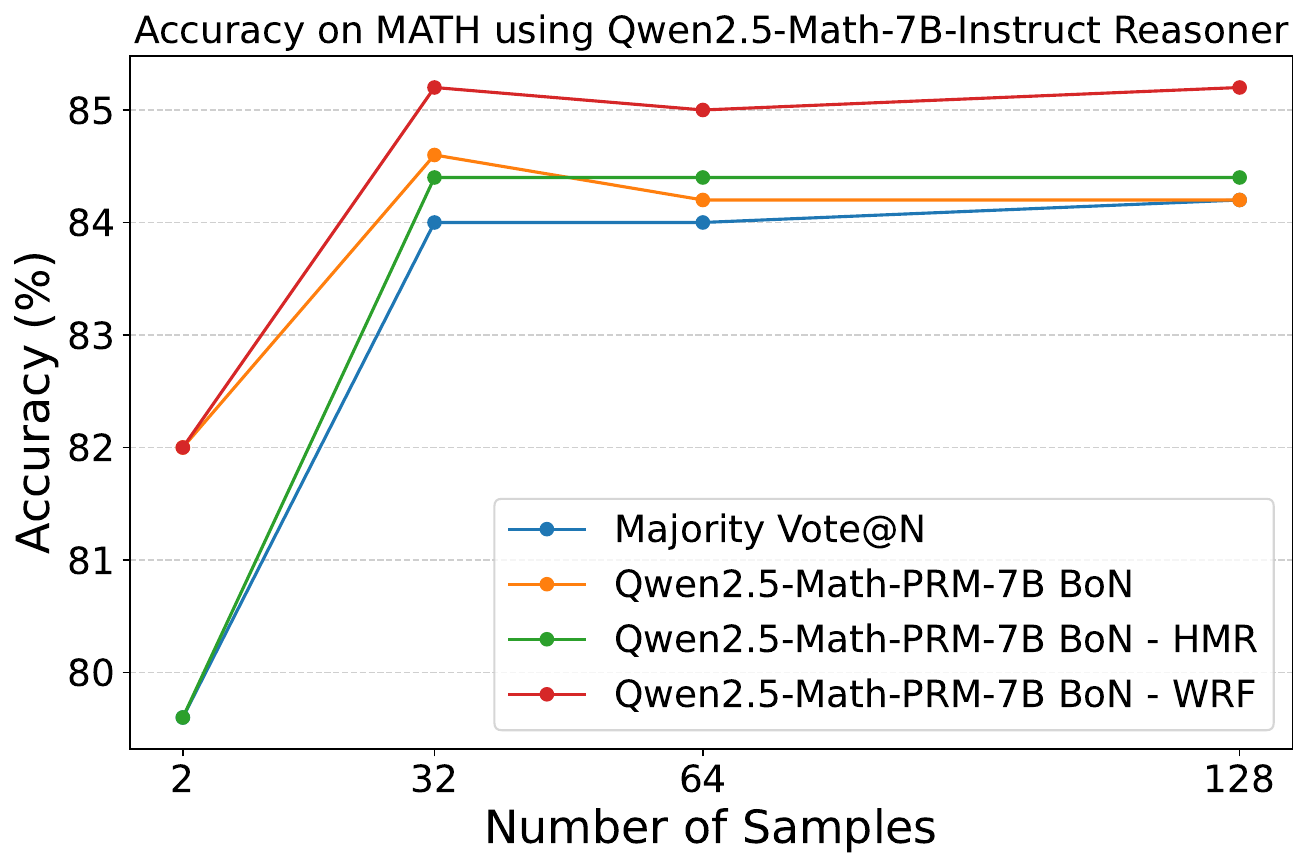}
  \end{subfigure}
  \begin{subfigure}[b]{0.3\textwidth}
    \includegraphics[width=\textwidth]{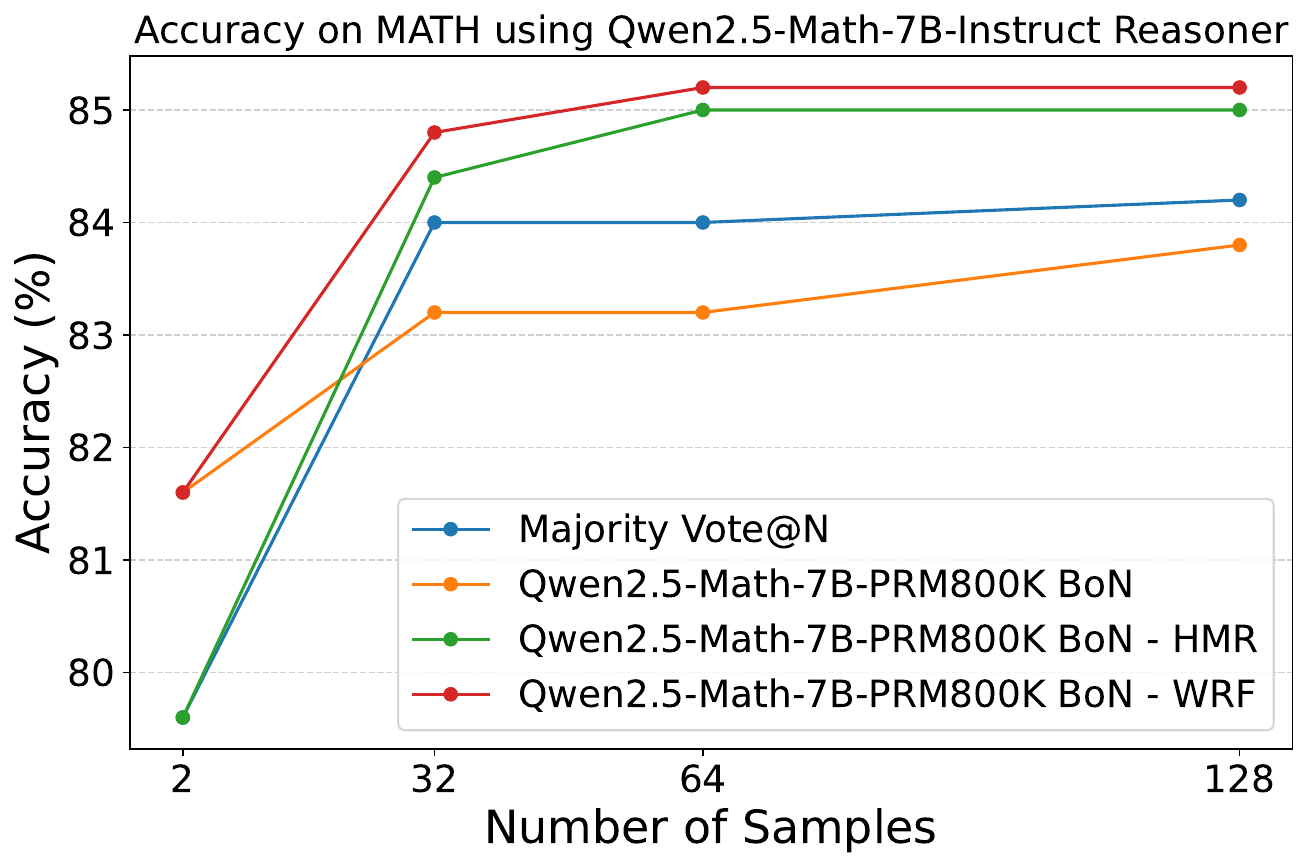}
  \end{subfigure}
  \\ 
  \begin{subfigure}[b]{0.3\textwidth}
    \includegraphics[width=\textwidth]{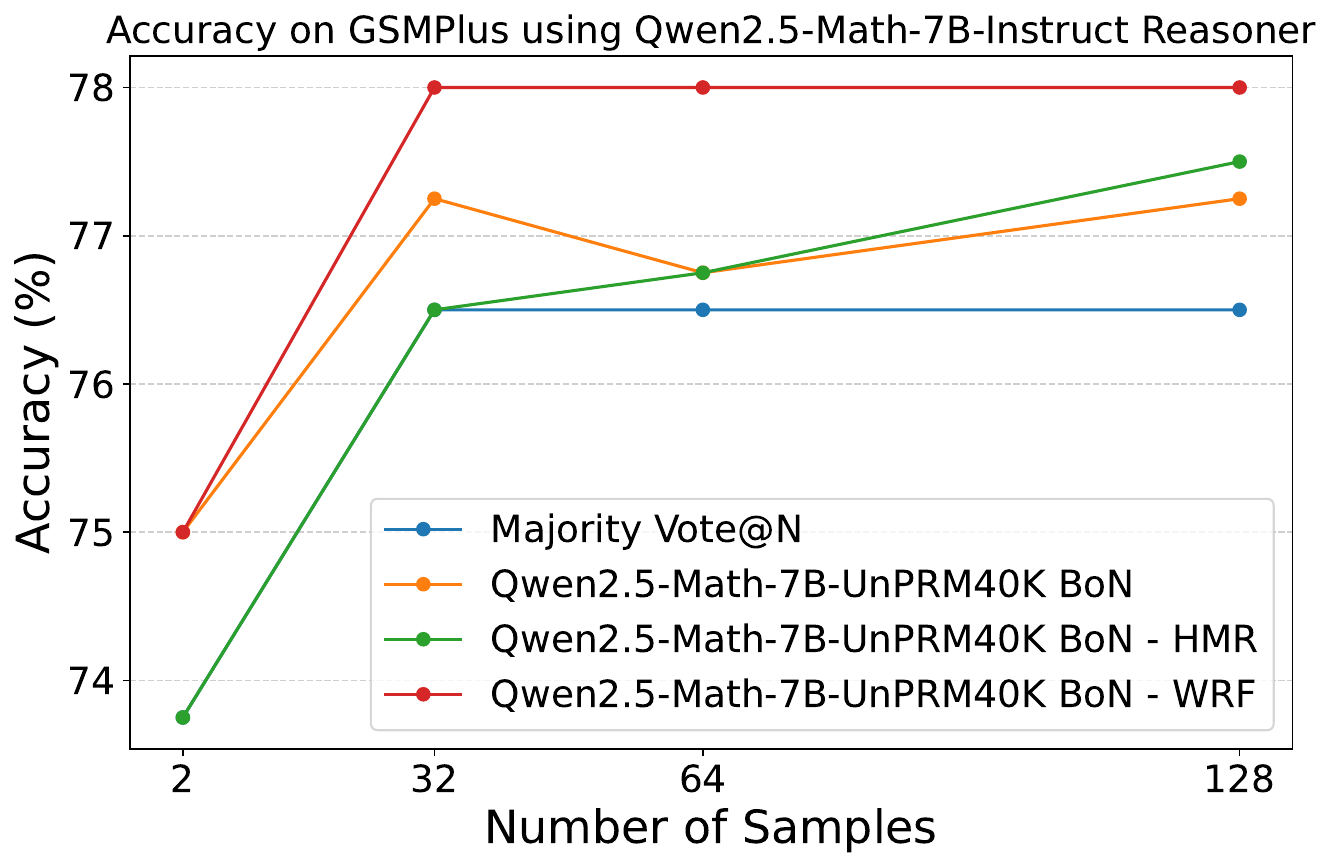}
  \end{subfigure}
  \begin{subfigure}[b]{0.3\textwidth}
    \includegraphics[width=\textwidth]{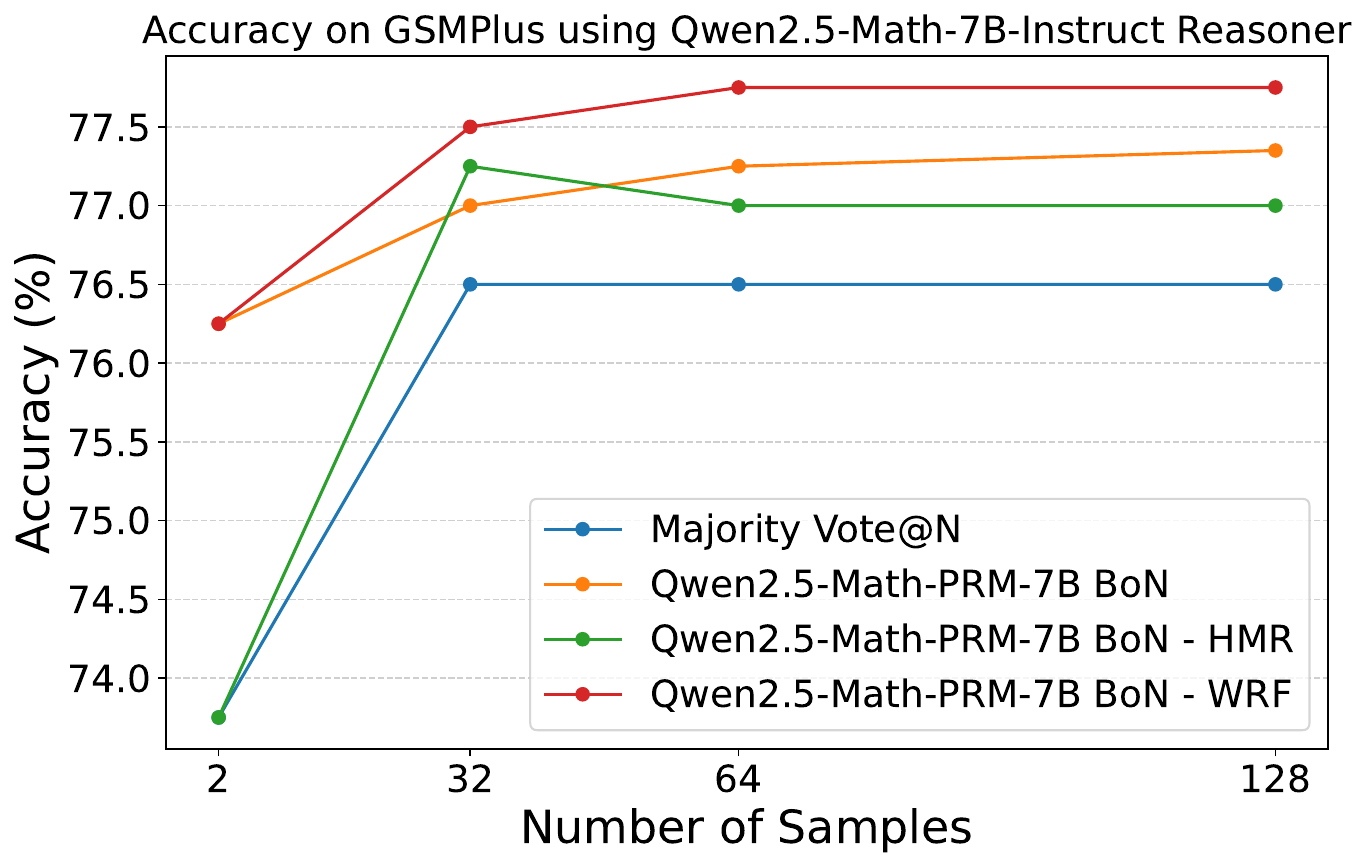}
  \end{subfigure}
  \begin{subfigure}[b]{0.3\textwidth}
    \includegraphics[width=\textwidth]{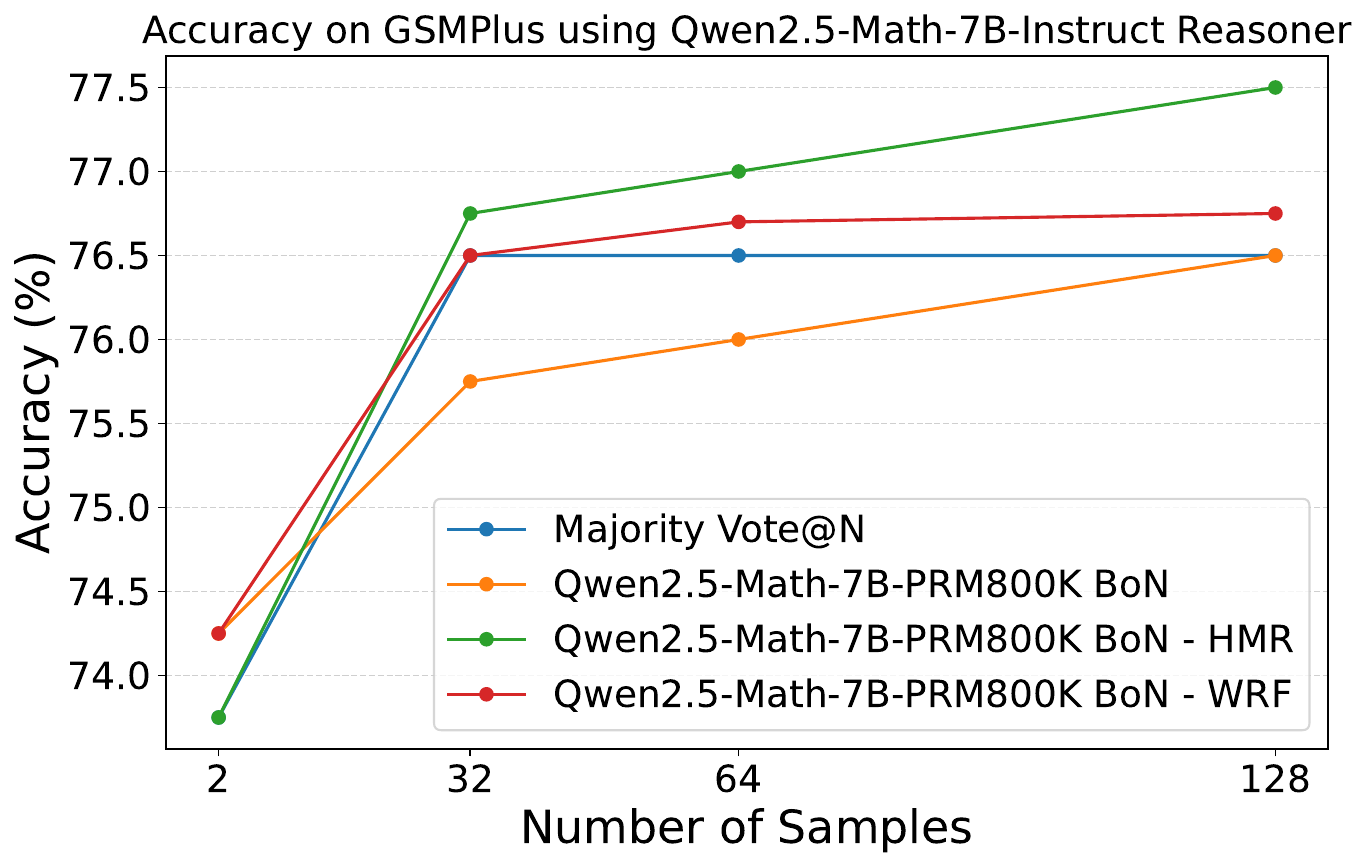}
  \end{subfigure}
  \caption{Evaluation results of different PRMs using diverse output aggregation strategies on MATH and GSMPlus datasets.}
\label{fig:hmr_wrf_qwen_results}
\end{figure*}

\subsubsection{Models and Baselines}

We utilise two models as the reasoner: the black-box LLM GPT-4o and the open-source LLM Qwen2.5-Math-7B-Instruct. For baseline PRMs, we compare our UnPRM40K with SimPRM40K, EpicPRM40K, and RanPRM40K. In addition to PRMs trained on datasets of equal size, we also compare UnPRM40K with two publicly available PRMs~\cite{DBLP:journals/corr/abs-2501-07301}: Qwen2.5-Math-PRM-7B and Qwen2.5-Math-7B-PRM800K, which are trained on 1.8M and 264K examples, respectively, substantially more than the 40K examples used for UnPRM40K. For output aggregation methods, we compare our proposed HMR vote and WRF vote strategies with standard PRM vote and standard Majority Vote.

\subsubsection{Results}

Figure~\ref{fig:40kprms_results} shows the accuracy of various PRM-based BoN aggregation strategies, alongside Majority Vote, on the MATH and GSMPlus datasets using both GPT-4o and Qwen2.5-Math-7B-Instruct reasoners. Across all configurations, UnPRM40K consistently outperforms SimPRM40K, demonstrating that uncertainty-driven PRM data generation is more effective than the similarity-driven approach. UnPRM40K also performs comparably to EpicPRM40K, which annotates incorrect solutions by identifying the first erroneous step, thereby validating the efficiency and effectiveness of our uncertainty-driven annotation method that locates the most uncertain error. As expected, RanPRM40K performs the worst, however, it still shows some improvement due to the correct labelling of correct solutions. 

Figure~\ref{fig:hmr_wrf_qwen_results} presents the results of various PRMs combined with different output aggregation strategies on the MATH and GSMPlus datasets, using the Qwen2.5-Math-7B-Instruct reasoner. The findings show that, across all PRMs, both the WRF and HMR strategies consistently outperform standard Majority Vote and traditional PRM-based methods. Performance increases with the number of samples in every setting. Notably, when standard PRM methods underperform Majority Vote, employing HMR and WRF leads to substantial performance gains. Among the two uncertainty-aware aggregation strategies, WRF demonstrates greater robustness than HMR in most scenarios. These results highlight the effectiveness of uncertainty-aware output aggregation methods that integrate the complementary strengths of Majority Vote and PRMs. Notably, applying the WRF strategy to UnPRM40K yields the best performance among the three PRMs, indicating that our uncertainty-driven data generation approach is particularly well-suited for enhancing uncertainty-aware aggregation methods like WRF vote. Due to the page limit, we put the GPT-4o reasoner results in Appendix.
\begin{table}[t]
\centering
\resizebox{\columnwidth}{!}{%
\begin{tabular}{@{}lcccccc@{}}
\toprule
\textbf{Model} & \rotatebox{45}{\textbf{Training Data Size}} & \rotatebox{45}{\textbf{GSM8K}} & \rotatebox{45}{\textbf{MATH}} & \rotatebox{45}{\textbf{Olympiad-Bench}} & \rotatebox{45}{\textbf{Omni-MATH}} & \rotatebox{45}{\textbf{Average}} \\\midrule
Math-Shepherd-PRM-7B       & 445K   & 47.9 & 29.5 & 24.8 & 23.8 & 31.5 \\
RLHFlow-PRM-Mistral-8B     & 273K   & 50.4 & 33.4 & 13.8 & 15.8 & 28.4 \\
RLHFlow-PRM-Deepseek-8B    & 253K   & 38.8 & 33.8 & 16.9 & 16.9 & 26.6 \\
EurusPRM-Stage2-7B         & 500K   & 47.3 & 35.7 & 21.2 & 20.9 & 31.3 \\ \midrule
Qwen2.5-Math-7B-RanPRM40K  & 40K    & 35.5 & 25.5 & 15.7 & 17.3 & 23.5 \\
Qwen2.5-Math-7B-SimPRM40K  & 40K    & 51.2 & 38.5 & 29.5 & 27.4 & 36.7 \\
Qwen2.5-Math-7B-UnPRM40K   & 40K    & 53.5 & 43.4 & 33.6 & 30.8 & 40.3 \\
Qwen2.5-Math-7B-EpicPRM40K & 40K    & 53.1 & 44.6 & 31.8 & 33.6 & 40.7 \\ \midrule
Qwen2.5-Math-7B-PRM800K    & 264K   & 68.2 & 62.6 & 50.7 & 44.3 & 56.5 \\
Qwen2.5-Math-PRM-7B        & 1.8M   & 82.4 & 77.6 & 67.5 & 66.3 & 73.5 \\
\bottomrule
\end{tabular}}
\caption{Evaluation results on ProcessBench. We report the F1 score of the respective accuracies on erroneous and correct samples. Among these PRMs, only Qwen2.5-Math-7B-PRM800K is trained on the human annotation data.}
\label{tab:processbench_result}
\end{table}
\subsection{PRM ProcessBench Evaluation}
ProcessBench ~\cite{DBLP:journals/corr/abs-2412-06559} is a benchmark to measure the ability to identify erroneous steps in mathematical reasoning. It consists of $3{,}400$ test cases, primarily focused on competition-level math problems. We test PRMs on it evaluate the step-level process errors identification ability.

\noindent\textbf{Baselines.} In addition to the above mentioned PRMs, we also compare with the following PRMs: Math-Shepherd-PRM-7B~\cite{DBLP:conf/acl/WangLSXDLCWS24}, RLHFlow-PRM-Mistral-8B~\cite{DBLP:journals/tmlr/Dong0000Z0SX024}, RLHFlow-PRM-Deepseek-8B~\cite{DBLP:journals/tmlr/Dong0000Z0SX024}, EurusPRM-Stage2-7B~\cite{DBLP:journals/corr/abs-2502-01456}.

\noindent\textbf{Results.} The evaluation results on ProcessBench are presented in Table~\ref{tab:processbench_result}. UnPRM40k outperforms PRMs trained on automatically labelled datasets containing $200K–500K$ examples. In comparison to SimPRM40k, the results indicate that PRMs trained on similarity-driven data are less effective than those trained on uncertainty-driven data. As expected, RanPRM40k demonstrates limited ability to identify erroneous steps, resulting in a low F1 score, while EpicPRM40k, which is annotated using the first error step, performs slightly better than UnPRM40k. Qwen2.5-Math-7B-PRM800K, which benefits from high-quality human-annotated data, achieves strong performance with only $264K$ training examples, illustrating that data quality can substantially improve PRM performance. Scaling the training set size to 1.8M in Qwen2.5-Math-PRM-7B further boosts the F1 score on ProcessBench, highlighting the effectiveness of data scaling in PRM training. 

\section{Analysis}
\begin{table}[t]
\centering
\resizebox{\columnwidth}{!}{%
\begin{tabular}{@{}lccc@{}}
\toprule
\textbf{Algorithm}                        & \textbf{Verified Steps}  & \textbf{Sampled Num.}    & \textbf{Generated Tok.}  \\ \midrule
Adaptive Binary Search                    & 3144                     & 104.98K                    & 36.44M                     \\
\textbf{Ours} & \textbf{1498 (-52\%)} & \textbf{69.25K (-34\%)} & \textbf{21.75M (-40\%)} \\ \bottomrule
\end{tabular}}
\caption{Computational cost of two automated PRM data annotation algorithms when annotating the same 1500 solutions (460 correct solutions, 1040 incorrect solutions).}
\label{tab:cost_analysis}
\end{table}
\noindent\textbf{Computational Cost Analysis.} Table~\ref{tab:cost_analysis} presents the computational cost analysis for two automated PRM data annotation algorithms applied to the same set of $1{,}500$ solutions, comprising $460$ correct and $1{,}040$ incorrect cases. Annotation of correct solutions does not require any sampling, so the computational cost is primarily driven by the annotation of incorrect solutions. The Adaptive Binary Search method~\cite{DBLP:journals/corr/abs-2503-02382}, used in our EpicPRM40K, annotates data by identifying the first erroneous step through a binary search process. In contrast, the Uncertainty-driven Search method, used in UnPRM40K and detailed in Algorithm~\ref{alg:uncertainty_process_annotation}, locates the most uncertain erroneous step for annotation. Both methods were run on a single A100 GPU. The results demonstrate that our approach substantially reduces the number of verified steps, sampled instances, and generated tokens compared to the Adaptive Binary Search method. Not only is the Uncertainty-driven Search more cost-effective, but it also achieves comparable performance to the Adaptive Binary Search, as shown in Figure~\ref{fig:40kprms_results}.



\begin{figure}[t]
    \centering
    \begin{subfigure}[b]{0.48\textwidth}
        \centering
        \includegraphics[width=\textwidth]{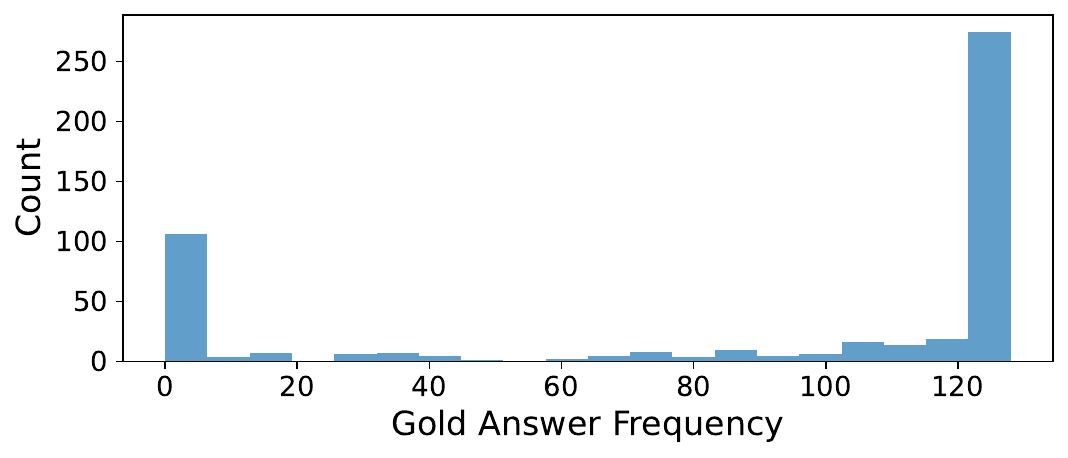}
        \caption{Visualisation of the distribution of gold answer frequencies in 128 outputs sampled from Qwen2.5-Math-7B-Instruct on MATH.}
        \label{fig:count_frequency_visual}
    \end{subfigure}
    \hfill
    \begin{subfigure}[b]{0.48\textwidth}
        \centering
        \includegraphics[width=\textwidth]{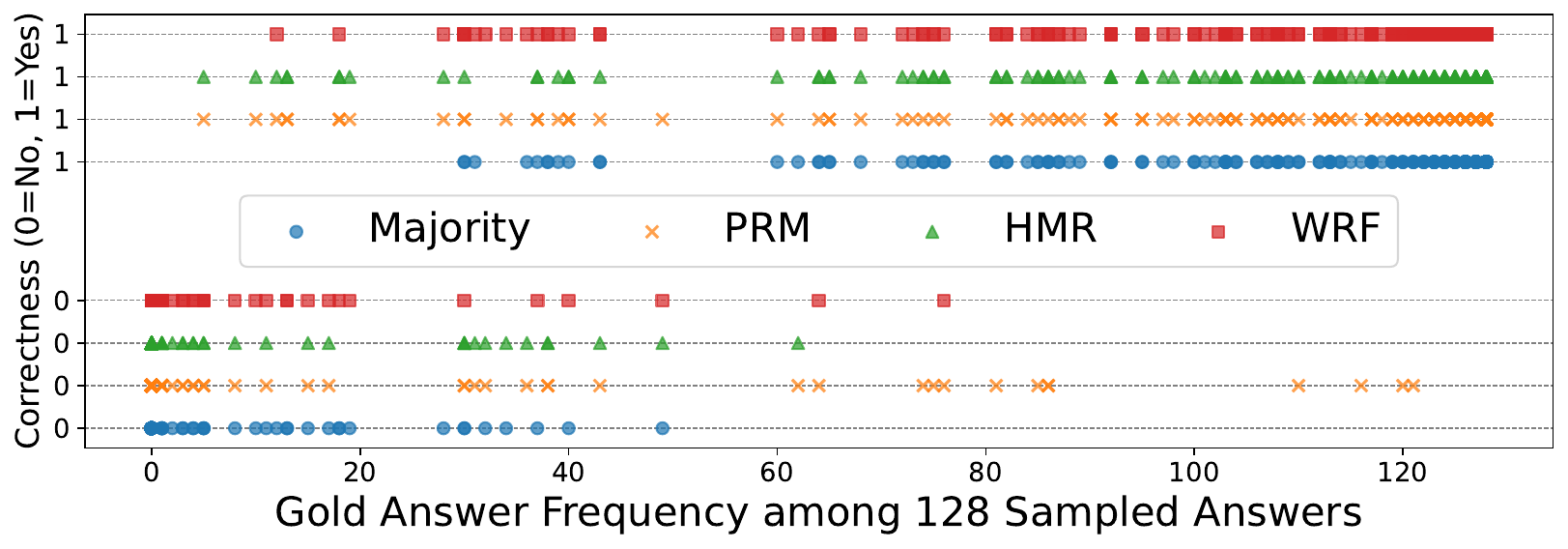}
        \caption{Visualisation of correctness for four output aggregation methods (Majority, PRM, HMR, WRF) over 128 outputs across gold answer frequencies on MATH. Each method is shown in a separate lane, with points indicating whether predictions are correct (1) or incorrect (0) at each frequency. The reasoner is Qwen2.5-Math-7B-Instruct, and the PRM used is Qwen2.5-Math-7B-UnPRM40K.}
        \label{fig:answer_frequency_visual}
    \end{subfigure}
    \caption{(a) Distribution of gold answer frequencies and (b) correctness visualisation for aggregation methods.}
    \label{fig:answer_and_count_frequency_visual}
\end{figure}
\noindent\textbf{Output Aggregation Visualisation.} Figure~\ref{fig:count_frequency_visual} displays the distribution of gold answer frequencies in 128 outputs sampled from Qwen2.5-Math-7B-Instruct on the MATH dataset. Among the 500 math questions, for more than half, the model consistently predicts the correct answer in all 128 samples. Conversely, for roughly 100 questions, the model fails to generate the correct answer even once within the 128 outputs. These results highlight a wide range of frequency distributions, suggesting a substantial proportion of questions where answer selection is non-trivial. In these cases, PRM-based methods are expected to help.

Figure~\ref{fig:answer_frequency_visual} illustrates the accuracy of four output aggregation methods: Majority, PRM, HMR, and WRF—across varying gold answer frequencies. When the gold answer appears with high frequency (over 60 times), the Majority method reliably selects the correct answer, indicating strong model confidence. However, its performance deteriorates when the gold answer frequency drops below 20, often failing to recover the correct response. In contrast, the PRM method can still identify some correct answers even when the gold answer is infrequent (below 20), though it may make mistakes in high-frequency scenarios where the Majority method succeeds. The HMR and WRF strategies, which integrate both Majority and PRM signals, notably reduce errors in the high-frequency regime where PRM alone underperforms. Moreover, WRF outperforms HMR in the mid-frequency range (between 20 and 40), yielding more correct predictions. These findings demonstrate the effectiveness of the two proposed uncertainty-aware aggregation methods in leveraging both model consensus and reasoning confidence to improve answer selection.

\noindent\textbf{Error Step Uncertainty Analysis.} UnPRM40k was generated and annotated using 3 different LLMs, with the dataset statistics summarised in Table~\ref{tab:data_stat}. The average number of sampled steps reflects how many steps the uncertainty-driven search algorithm must verify on average to locate the most uncertain erroneous step, where a value of 1 represents optimal efficiency. Across all three models, the results are very close to 1, indicating that the uncertainty-driven search algorithm is highly efficient in pinpointing the most uncertain error. The average error step uncertainty rank indicates the uncertainty rank of the identified erroneous step, with 0 as the optimal value. Again, results are consistently near 0 across the three models, demonstrating that uncertainty serves as an effective proxy for locating errors. These findings are consistent with the intuition that LLMs are more likely to make mistakes where output is less certain.

\begin{table}[t]
\centering
\resizebox{0.45\textwidth}{!}{%
\begin{tabular}{@{}lccc@{}}
\toprule
\textbf{Sampling Model}  & \textbf{\begin{tabular}[c]{@{}c@{}}Num of Samples\end{tabular}} & \textbf{\begin{tabular}[c]{@{}c@{}}Avg Num\\ Sample Step\end{tabular}} & \textbf{\begin{tabular}[c]{@{}c@{}}Avg Error Step \\ Uncertainty Rank\end{tabular}} \\ \midrule
Llama-3.1-8B-Instruct    & 20,264                                                                      & 1.33                                                                    & 0.33                                                                                    \\
Qwen2.5-7B-Instruct      & 12,019                                                                      & 1.51                                                                    & 0.51                                                                                    \\
Mistral-7B-Instruct-v0.3 & 8,223                                                                       & 1.34                                                                    & 0.35                                                                                    \\ \bottomrule
\end{tabular}}
\caption{Statistics of the annotated dataset UnPRM40K.}
\label{tab:data_stat}
\end{table}

\section{Conclusion}
This work presents an uncertainty-driven framework for constructing and annotating process reward data, addressing the key challenges in training effective PRMs for mathematical reasoning with LLMs. By leveraging uncertainty estimation, we efficiently generate and label high-quality step-level supervision data, greatly reducing annotation costs while maintaining performance. Furthermore, our proposed HMR and WRF aggregation strategies successfully combine the strengths of majority vote and PRM-based methods, leading to more robust and accurate answer selection. Extensive experiments demonstrate that our PRM data construction framework enhances both the efficiency and effectiveness of PRM training, and that the proposed output aggregation strategies are both effective and generalise well across different PRMs and mathematical reasoning tasks.

\section{Limitations}
While our uncertainty-aware aggregation methods incorporate answer frequency information, their performance can be influenced by the quality of the majority vote baseline. In scenarios where majority voting performs poorly, combining it with PRM-based approaches may not yield additional improvements and could potentially impact overall performance. However, we find that the proposed aggregation strategies are particularly effective when the majority vote performs better or comparable to the PRM.

\bibliography{aaai2026}

\appendix



\newpage

\section*{Appendix}
\label{sec:Appendix}

\subsection{Algorithms}
Hybrid Majority Reward Vote and Weighted Reward-Frequency Vote are demonstrated in Algorithm~\ref{alg:hmr_voting} and ~\ref{alg:wrf_voting}, respectively.

\subsection{Output Aggregation Results on GPT-4o}
\begin{figure*}[t]
  \centering
  \begin{subfigure}[b]{0.3\textwidth}
    \includegraphics[width=\textwidth]{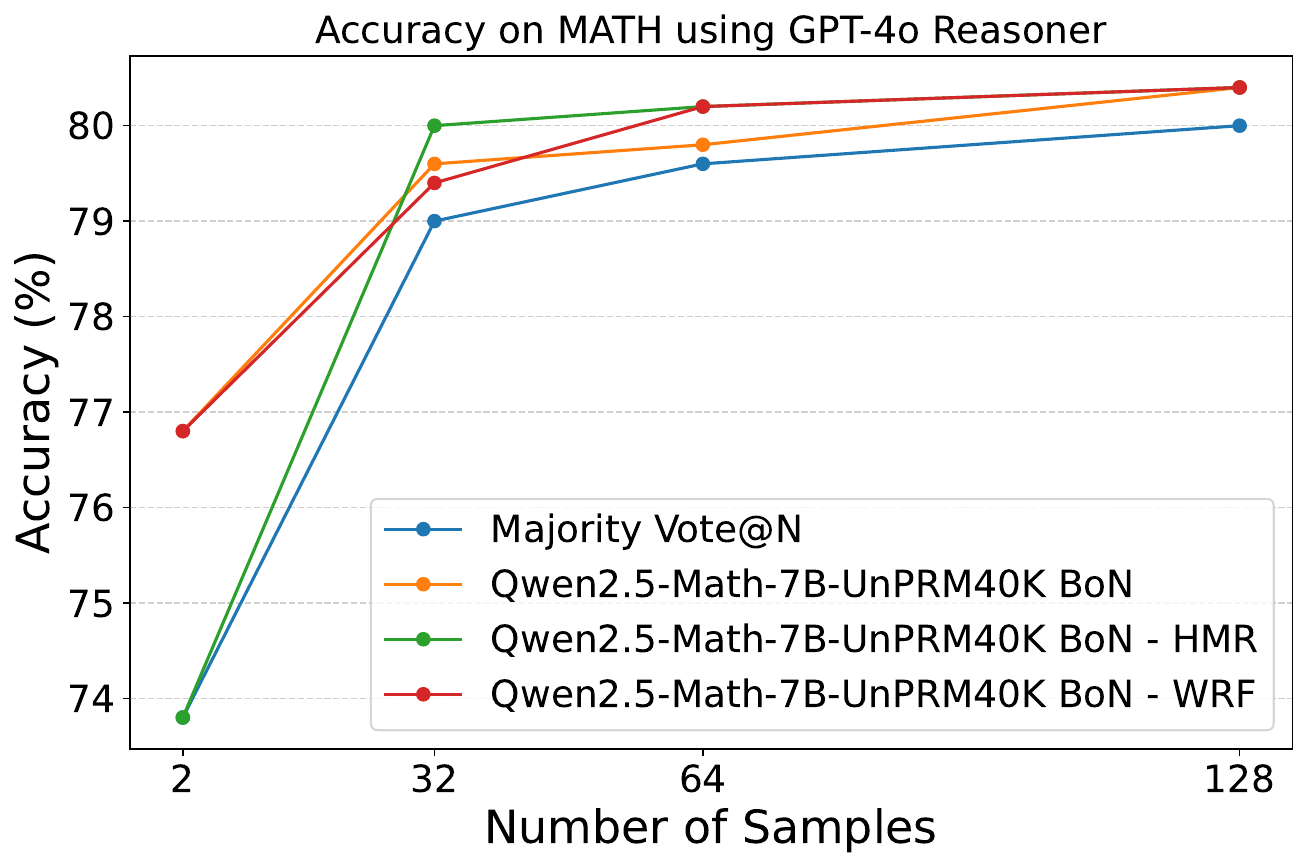}
  \end{subfigure}
  \begin{subfigure}[b]{0.3\textwidth}
    \includegraphics[width=\textwidth]{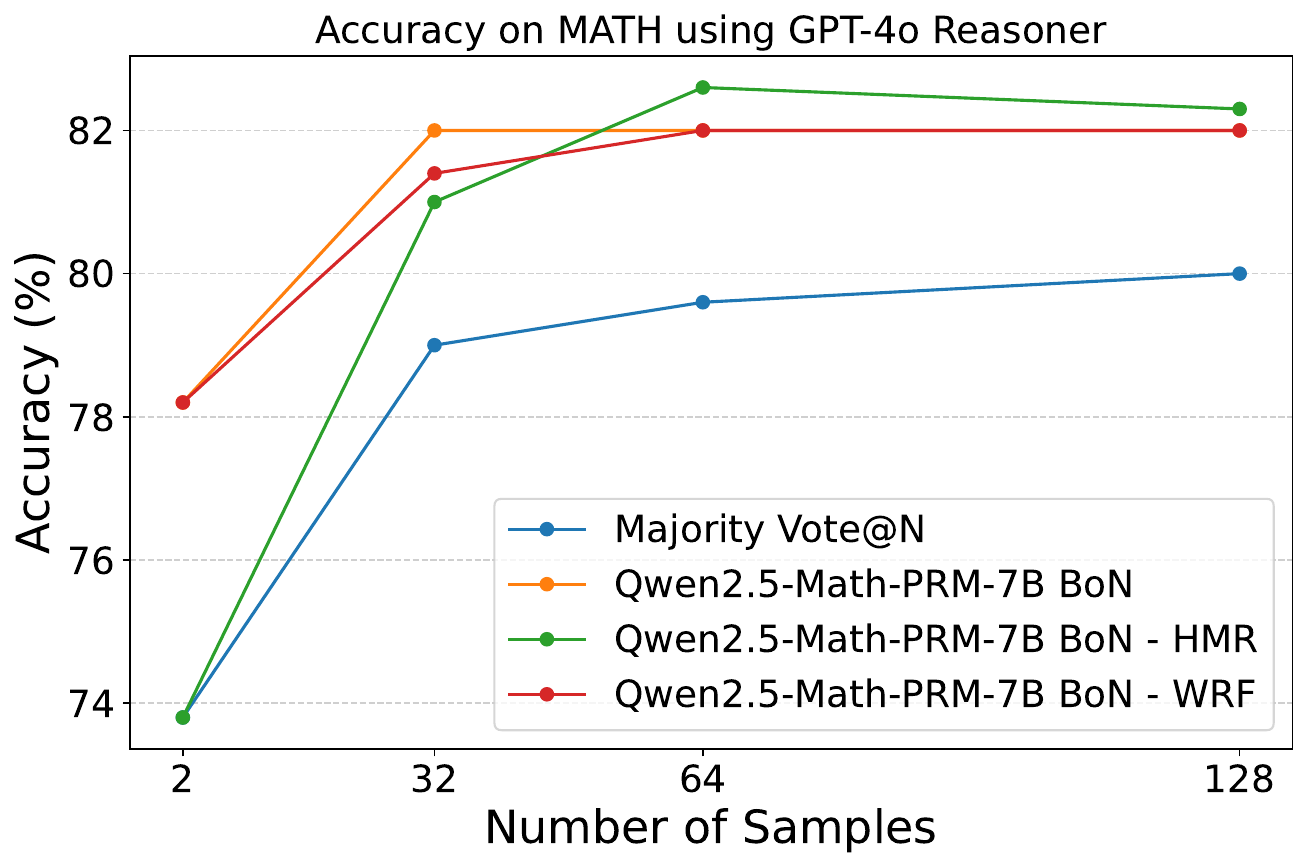}
  \end{subfigure}
  \begin{subfigure}[b]{0.3\textwidth}
    \includegraphics[width=\textwidth]{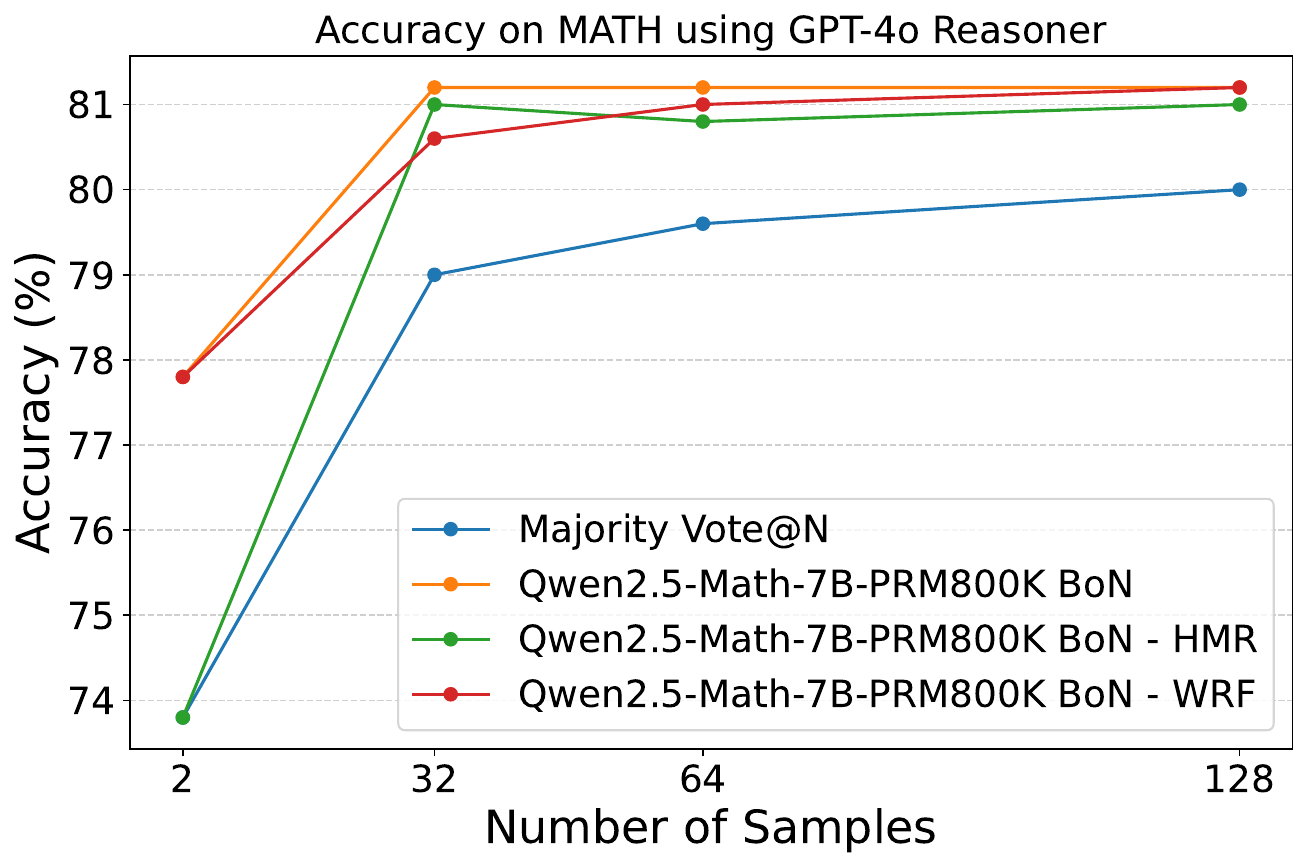}
  \end{subfigure}
  \\ 
  \begin{subfigure}[b]{0.3\textwidth}
    \includegraphics[width=\textwidth]{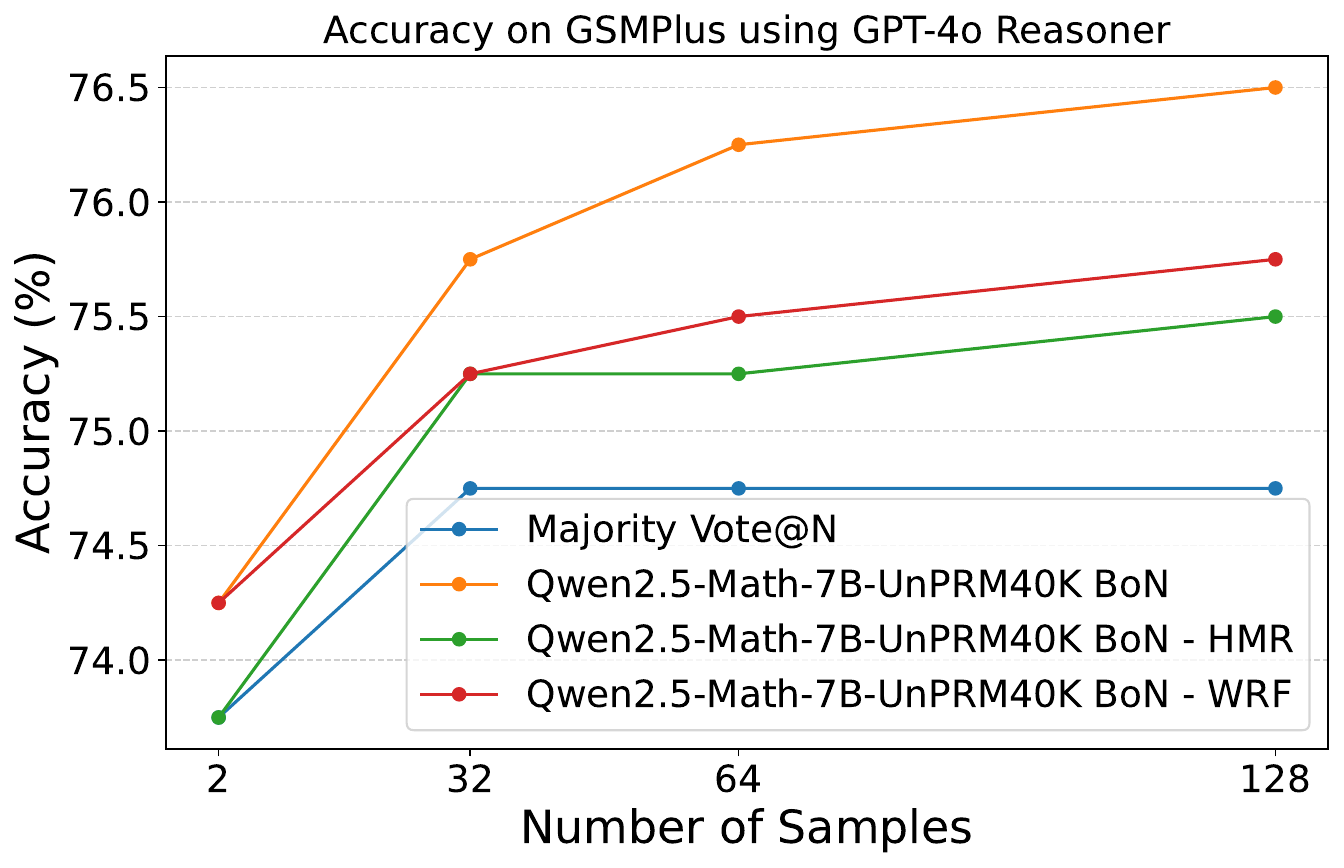}
  \end{subfigure}
  \begin{subfigure}[b]{0.3\textwidth}
    \includegraphics[width=\textwidth]{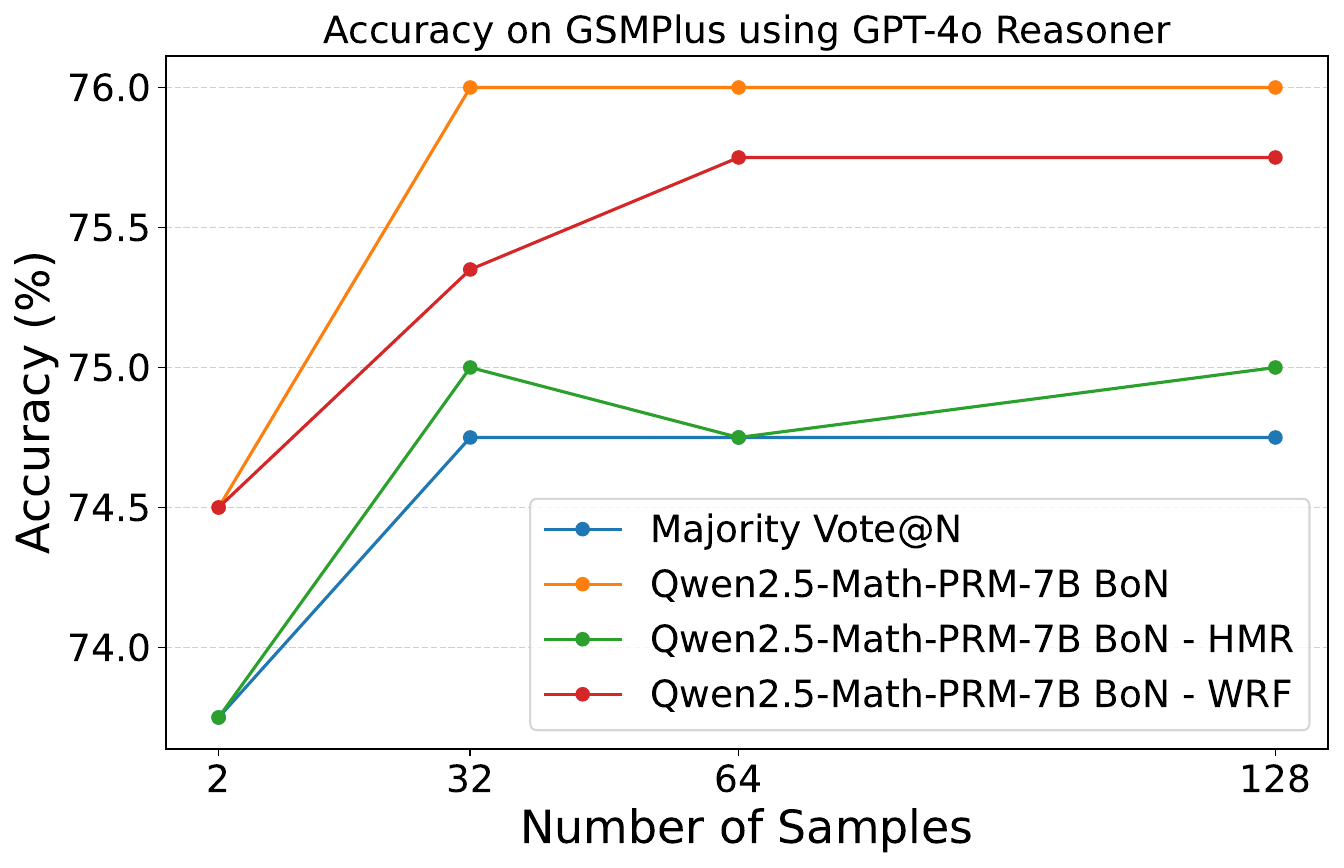}
  \end{subfigure}
  \begin{subfigure}[b]{0.3\textwidth}
    \includegraphics[width=\textwidth]{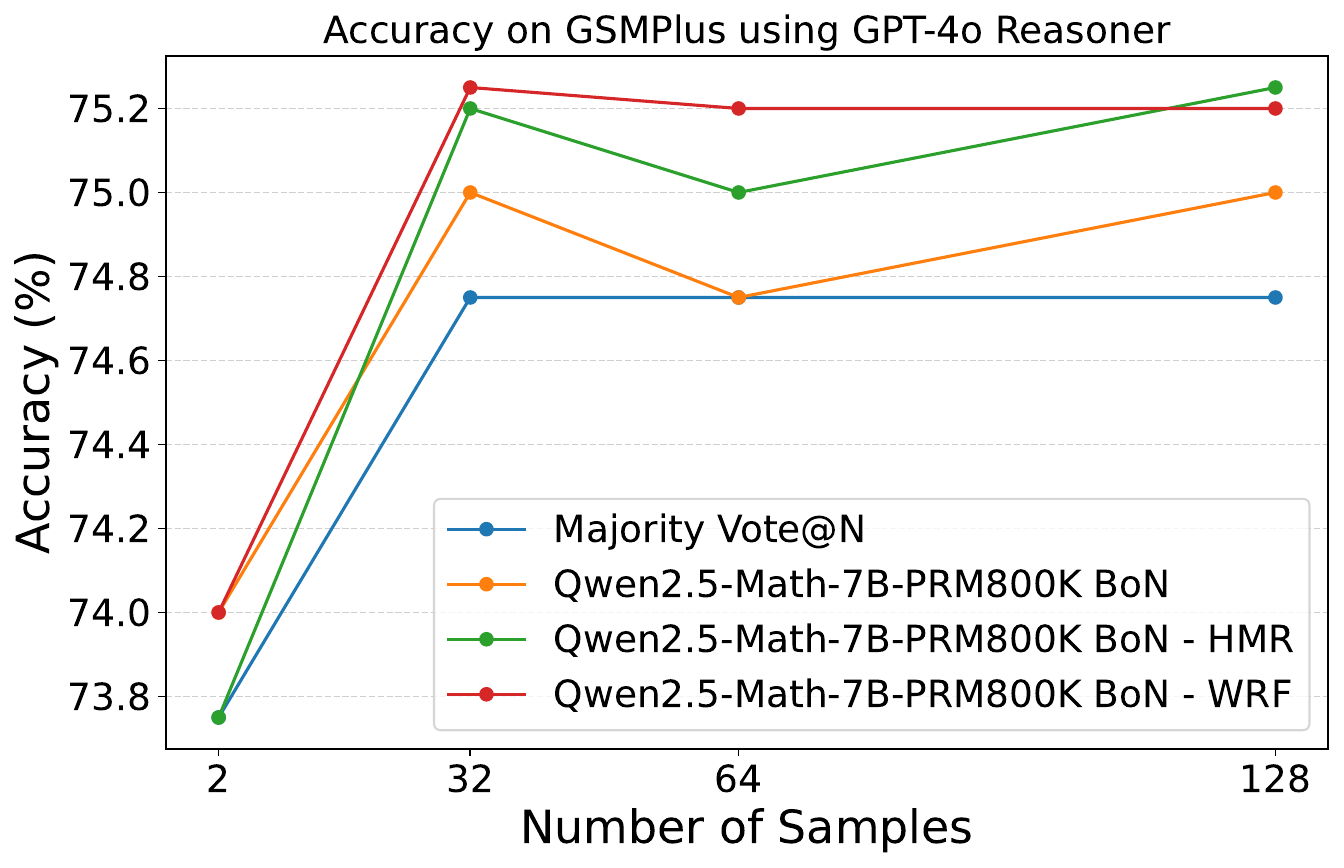}
  \end{subfigure}
  \caption{Evaluation results of different PRMs using diverse output aggregation strategies on two datasets with GPT-4o reasoner.}
\label{fig:hmr_wrf_gpt4o_results}
\end{figure*}

Figure~\ref{fig:hmr_wrf_gpt4o_results} presents the evaluation results for different PRMs using various output aggregation strategies on two datasets with the GPT-4o reasoner. On the MATH dataset (top three plots) and on GSMPlus with Qwen2.5-Math-7B-PRM800K (bottom right), uncertainty-aware vote strategies consistently outperform both Majority Vote and standard PRM Vote when the sample size is 128. However, in the two remaining GSMPlus settings (bottom left and centre), where Majority Vote significantly underperforms relative to standard PRMs, incorporating uncertainty-aware vote does not yield further improvements over the standard PRM result. This is because the poor performance of Majority Vote diminishes its contribution when combined with PRM-based methods. We discuss this finding in the Limitations section.

\begin{algorithm}[ht]
\caption{Hybrid Majority Reward (HMR) Vote}
\label{alg:hmr_voting}
\textbf{Input:} List of $N$ sampled candidate solutions $S = \{s_1, \ldots, s_N\}$ (each $s_i$ contains steps and a final answer); PRM function $R(\cdot)$ (returns list of solution step scores)\\
\textbf{Output:} Selected answer $a^*$

\begin{algorithmic}[1]
\STATE Extract answer $a_i$ from each solution $s_i$ in $S$, forming set $A = \{a_1, \ldots, a_N\}$
\STATE Count the frequency of each unique answer in $A$
\STATE Let $a_{\mathrm{maj}}$ be the answer with the highest frequency (majority vote)
\STATE Let $f_{\mathrm{maj}}$ be the frequency of $a_{\mathrm{maj}}$
\IF{$f_{\mathrm{maj}} < N/2$}
    \STATE \COMMENT{Majority is uncertain; use PRM to select the answer}
    \FOR{each solution $s_i$ in $S$}
        \STATE Compute step scores list: $R(s_i)$
        \STATE Compute solution reward: $r_i = \min R(s_i)$
    \ENDFOR
    \STATE Let $s^* = \arg\max_{s_i \in S} r_i$ \COMMENT{Select solution with the highest reward score}
    \STATE $a^* \gets$ final answer of $s^*$
\ELSE
    \STATE $a^* \gets a_{\mathrm{maj}}$ \COMMENT{Select majority vote answer}
\ENDIF
\STATE \textbf{return} $a^*$
\end{algorithmic}
\end{algorithm}

\begin{algorithm}[ht]
\caption{Weighted Reward-Frequency (WRF) Vote}
\label{alg:wrf_voting}
\textbf{Input:} List of $N$ sampled candidate solutions $S = \{s_1, \ldots, s_N\}$ (each $s_i$ contains steps and a final answer); PRM function $R(\cdot)$ (returns list of solution step scores)\\
\textbf{Parameter:} Weighting parameter $\alpha \in [0, 1]$ \\
\textbf{Output:} Selected answer $a^*$

\begin{algorithmic}[1]
\STATE Extract answer $a_i$ from each solution $s_i$ in $S$, forming set $A = \{a_1, \ldots, a_N\}$
\FOR{each solution $s_i$ in $S$}
    \STATE Compute step scores list: $R(s_i)$
    \STATE Compute solution reward: $r_i = \min R(s_i)$
\ENDFOR
\STATE Initialise dictionary $G$ mapping answer $a$ to list of rewards
\FOR{$i = 1$ to $N$}
    \STATE Append $r_i$ to $G[a_i]$
\ENDFOR
\FOR{each unique answer $a$ in $G$}
    \STATE Compute mean reward $m_a = \frac{1}{|G[a]|} \sum_{r \in G[a]} r$
    \STATE Compute frequency $f_a = |G[a]|$
\ENDFOR
\STATE Let $M_{\min} = \min\{ m_a \}$, $M_{\max} = \max\{ m_a \}$
\STATE Let $F_{\min} = \min\{ f_a \}$, $F_{\max} = \max\{ f_a \}$
\FOR{each unique answer $a$}
    \IF{$M_{\max} = M_{\min}$}
        \STATE $\hat{m}_a \gets 1.0$
    \ELSE
        \STATE $\hat{m}_a \gets \frac{m_a - M_{\min}}{M_{\max} - M_{\min}}$
    \ENDIF
    \IF{$F_{\max} = F_{\min}$}
        \STATE $\hat{f}_a \gets 1.0$
    \ELSE
        \STATE $\hat{f}_a \gets \frac{f_a - F_{\min}}{F_{\max} - F_{\min}}$
    \ENDIF
\ENDFOR
\FOR{each unique answer $a$}
    \STATE Compute combined score: $s_a = \alpha \cdot \hat{m}_a + (1-\alpha) \cdot \hat{f}_a$
\ENDFOR
\STATE $a^* = \arg\max_{a} s_a$
\STATE \textbf{return} $a^*$
\end{algorithmic}
\end{algorithm}



\end{document}